\documentclass[10pt,twocolumn,letterpaper]{article}

\usepackage{iccv}              

\usepackage{siunitx}
\usepackage{booktabs}
\usepackage{multicol}
\usepackage{multirow}
\usepackage{xcolor}
\usepackage{colortbl}
\usepackage{amsmath}
\usepackage{mathrsfs}
\usepackage{algorithm}
\usepackage{algorithmic}
\usepackage{xspace}
\definecolor{darkgreen}{rgb}{0.1, 0.5, 0.1}
\usepackage{listings}

\definecolor{iccvblue}{rgb}{0.21,0.49,0.74}
\usepackage[pagebackref,breaklinks,colorlinks,allcolors=iccvblue]{hyperref}

\def\paperID{2102} 
\def\confName{ICCV}
\def\confYear{2025}

\title{Beyond Intermediate States: 
Explaining Visual Redundancy through Language}

\author{Dingchen Yang\thanks{Work done during 
an internship at tencent hunyuan team}\\
Tongji University\\
{\tt\small dingchen\_yang@tongji.edu.cn}
\and
Bowen Cao\\
CUHK\\
\and
Anran Zhang\\
Tencent Hunyuan Team\\
\and
Weibo Gu\\
Tencent Hunyuan Team\\
\and
Winston Hu\\
Tencent Hunyuan Team\\
\and
Guang Chen\thanks{Corresponding author}\\
Tongji University\\
{\tt\small guangchen@tongji.edu.cn}
}

\begin{document}
\maketitle
\begin{abstract}
Multi-modal Large Langue Models 
(MLLMs)
often 
process
thousands of
visual tokens,
which
consume
a significant 
portion
of the context window
and impose
a substantial
computational burden.
Prior
work
has
empirically
explored
visual token pruning
methods
based on
MLLMs'
intermediate
states
(\eg, attention scores).
However, 
they
have limitations
in precisely
defining
visual
redundancy
due to
their
inability
to
capture
the 
influence
of
visual tokens
on
MLLMs’
visual understanding
(\ie, the
predicted
probabilities
for textual 
token candidates).
To 
address this issue,
we
manipulate
the visual input
and 
investigate
variations
in the textual output
from
both
token-centric
and 
context-centric
perspectives,
achieving
intuitive
and comprehensive
analysis.
Experimental results reveal that
visual tokens with 
low
ViT$-[cls]$ association
and
low
text-to-image
attention scores
can
contain
recognizable
visual cues
and 
significantly
contribute to
images' overall information.
To develop
a more
reliable
method
for
identifying
and
pruning
redundant
visual tokens,
we 
integrate
these two
perspectives
and
introduce
a 
context-independent condition
to 
identify
redundant
prototypes
from training images,
which 
probes
the redundancy
of each visual token
during inference.
Extensive experiments
on single-image,
multi-image
and video
comprehension
tasks
demonstrate
the
effectiveness
of
our method,
notably
achieving
90\% 
to
110\%
of the
performance
while 
pruning 80\% to 90\% of
visual tokens.
Code will be available
at \url{https://github.com/DingchenYang99/RedundancyCodebook.git}.
\end{abstract}
\section{Introduction}
\label{sec:intro}

\begin{figure*}[tb]
  \centering
  \includegraphics[width=0.9\linewidth]{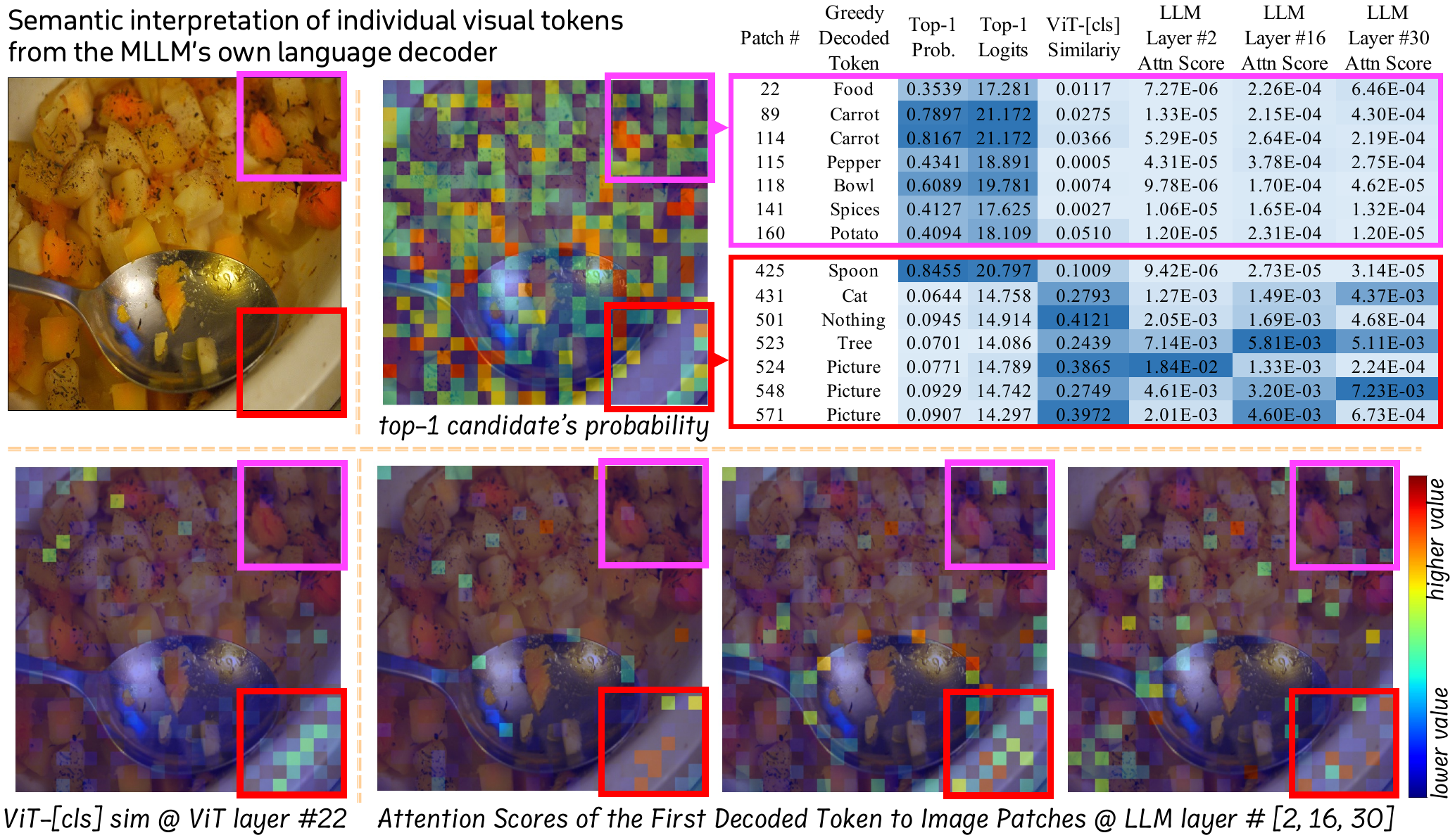}
  \caption{
    We 
    investigate the inherent information
    encoded in individual visual tokens
    by instructing 
    LLaVA-Next
    to describe
    them
    and analyzing
    the corresponding
    decoding results,
    predicted
    probabilities,
    and
    confidence scores
    (logits).
    ``Patch \#''
    indicates the index
    in the flattened
    patch sequence.
    Some visual tokens
    with
    low ViT$-[cls]$ similarity
    and low attention scores
    (\eg, Patch \#114, \#160, and \#425)
    contain valid visual information
    (\eg, \textit{Carrot},
    \textit{Potato},
    and \textit{Spoon})
    that
    the model recognizes with high confidence
    (40\% to 80\%
    probability).
    Conversely, 
    despite having
    high ViT$-[cls]$ similarity
    and high attention scores
    (highlighted in the red box),
    certain visual tokens yield
    text descriptions
    unrelated to the image patches
    (\eg, \textit{Cat} and \textit{Tree}),
    with model confidence lower than 10\%.
  }
  \label{fig:analysis_compare}
\end{figure*}

Multi-modal Large Language Models 
(MLLMs)~\cite{liu2024llavanext,li2024llavanext-strong,li2024llavaov}
have 
demonstrated
remarkable performance
across 
a range of
vision-language tasks,
including
high-resolution
image
and
video comprehension, 
by
integrating
thousands
of
visual tokens.
However,
This 
approach
introduces several challenges.
First,
visual tokens
encroach upon
the context window
required for textual tokens, 
and may interfere with 
MLLMs'
text processing 
capabilities~\cite{yang2024enhancing}.
Second,
the quadratic complexity of the
self-attention 
mechanism~\cite{vaswani2017attention}
significantly 
increases
the computational burden. 
Consequently,
reducing
redundant
visual tokens
is crucial for
enhancing 
the
overall performance
and efficiency
of MLLMs.

To 
reduce the number of visual tokens
while mitigating performance degradation,
recent research has empirically explored leveraging
MLLMs'
\textit{intermediate states}
to guide
inference-time visual token pruning.
The two primary approaches are:
(1) utilizing the
ViT$-[cls]$ 
token~\cite{shang2024llavaprumerge},
which encodes
global image information,
and (2) 
leveraging the
scalar attention scores
of textual tokens to visual tokens
in the LLM~\cite{chen2025fastv},
which capture cross-modal information flow.
However,
these
intermediate-state-based
methods
struggle to
explicitly characterize
the influence of
each visual token
on MLLMs'
visual understanding outcome,
\ie, the final probability prediction,
as
attention value vectors also
play a crucial role
in the attention mechanism,
and
the representation of
one token
progressively transforms
into that of
its next token
in auto-regressive LLMs.
This limitation hinders
the interpretable definition of
visual redundancy
and risks pruning
informative visual tokens
in MLLMs.

In this study,
we aim to provide a more precise explanation
of visual redundancy in MLLMs,
which first requires identifying
the direct impact of each visual token
on MLLMs' visual understanding.
Since humans
understand images
by
attending to
individual visual cues
and assessing their
contributions
to the overall image representation,
we analyze the influence of visual tokens from
two
perspectives:
\textbf{(1)
\textit{Token-Centric perspective}},
which 
examines the inherent visual information
encoded in each visual token,
and 
\textbf{(2)
\textit{Context-Centric perspective}},
which evaluates how
each visual token affects
the broader visual context
(\ie, images or image regions).
For the 
\textit{token-centric} analysis,
we
devise a 
\textit{single-token-input}
experiment
(\Cref{sec:single_input}),
isolating
each
visual token
and
instructing the MLLM
to interpret
the 
information
it contains.
This experiment
reveals that
MLLM can recognize
valid
visual information
from
visual tokens with low ViT$-[cls]$ similarity
and low text-to-image attention scores.
As shown in~\Cref{fig:analysis_compare},
LLaVA-Next~\cite{li2024llavanext-strong}
predicts
\textit{Carrot} and \textit{Spoon}
with over 80\% confidence
from patches
\#114 and \#425,
which depict
carrot and spoon,
respectively.
For the
\textit{context-centric} analysis,
we design a
\textit{leave-one-token-out}
experiment
(\Cref{sec:leave_one_out})
to examine how
removing individual visual tokens
affects the predicted probability distribution.
Experimental results
indicate that
certain
visual tokens
with
low ViT$-[cls]$ similarity
and low text-to-image attention scores
can still
significantly
influence
MLLMs' understanding
of their associated image
(\Cref{sec:discovery}).
These findings
warrant a reconsideration of
the definition of visual redundancy in MLLMs.

Based on our
token-centric and context-centric analyses,
we propose that 
redundant visual tokens
should be identified
according to two fundamental criteria:
the visual token
(1) lacks
recognizable
visual information
and
(2) does not
significantly
impact the overall information
of its associated image.
Building on the feature analysis
of visual tokens that satisfy these criteria,
we introduce a 
\textit{context-independent condition}
to
identify
\textit{redundant prototypes}
that are unlikely to
influence visual information across different images,
thus demonstrating potential for generalization.
Leveraging these criteria,
we propose an identify-then-probe strategy
for inference-time visual token pruning.
First,
We use training images
to identify
\textit{redundant prototypes}
and store them in an extensible 
\textit{redundancy codebook}.
During inference,
visual tokens
that exhibit
higher similarity
to these
prototypes
are 
deemed
more likely to be redundant
and are removed
before
sending
to
the LLM.

We evaluate
the effectiveness
of our approach
on five single-image
Visual Question Answering
(VQA) 
benchmarks~\cite{li2023evaluating,fu2024mme,liu2025mmbench,li2023seed,rwqa}
and two image captioning 
benchmarks~\cite{agrawal2019nocaps,flickr30k}.
On average, our 
method
preserves 90\%
of the performance of
LLaVA-Next~\cite{liu2024llavanext,li2024llavanext-strong}
and
LLaVA-1.5~\cite{liu2023llava,liu2024improved}
while pruning 90\% of visual tokens,
outperforming
representative
methods~\cite{shang2024llavaprumerge,chen2025fastv}
that
rely on
MLLMs'
\textit{intermediate states}.
Furthermore,
our approach is adaptable to
multi-image and video comprehension 
tasks~\cite{jiang2024mantis,wang2024muirbench,li2024mvbench},
achieving up to a 10\%
performance improvement for 
LLaVA-OneVision~\cite{li2024llavaov}
while
pruning
80\%
of
visual tokens.
These 
results
validate
the effectiveness
of our approach
\section{Related Work}
\label{sec:related_work}

Leading 
MLLMs~\cite{liu2024improved,liu2024llavanext,li2024llavaov,li2024llavanext-strong}
process
high-resolution images
and multiple video frames
by
incorporating numerous
visual tokens.
For instance,
LLaVA-Next
and LLaVA-OneVision
represent an image
using a maximum of
2,880 and 7,290 visual tokens, 
respectively.
These visual tokens
occupy a 
large
portion
of the context window of 
their LLM\footnote{Length of
8,192 
tokens
for
LLaMA3~\cite{llama3modelcard}
and
32,768
for
Qwen2~\cite{qwen2}.},
leading to
increased
computational overhead
and 
potentially
impairing MLLMs' 
text processing
capabilities~\cite{yang2024enhancing}.

\subsection{Identifying
Redundant Visual Tokens}
To 
alleviate
the computational burden
associated with
visual tokens,
pioneering
studies
explore
MLLMs'
\textit{intermediate states}
to estimate
the redundancy
of visual tokens.
The methodologies
can be 
broadly 
categorized
into 
two types: 

\subsubsection{Vision-Centric
Visual Redundancy Estimation}
This line of work
presumes
that
visual tokens
that do not
align with
the
overall
information
of the image
or 
exhibit
duplicated features
are redundant.
The alignment
between
image patches
and the image's
overall
information
is 
evaluated
by their
association
with
the $[cls]$ token
in the Vision Transformer
(ViT~\cite{dosovitskiy2020image})
model~\cite{shang2024llavaprumerge,zhang2024token,yang2024enhancing,liu2024multistage,han2024rethinking},
or by the attention scores
between one image patch
and all other
patches~\cite{xu2024freepruner,yang2024visionzip,zhang2024cls,wang2024cls}.
To identify
duplicate
visual tokens,
the feature similarity
of
patches
within a local spatial region~\cite{zhang2024token,liu2024multistage}
or a spatio-temporal region~\cite{shen2024longvu,tao2024dycoke}
is
assessed.
These strategies
typically 
distinguish
foreground objects
from
background patches.
However, 
given that
visual tokens 
are further processed
by the LLM
during the 
\textit{prefill stage}\footnote{The first forward
computation process
in the LLM
that decode the first token
utilizing all visual and
textual token embeddings.}
for
cross-modal feature interaction
and 
text decoding,
we advocate 
for explaining the 
information
encoded in visual tokens
from the 
viewpoint
of the LLM.

\subsubsection{Instruction-Based
Visual Redundancy Estimation}
This line of work
focuses on
the
cross-modal 
information flow
within LLMs,
identifying
visual tokens
that are irrelevant
to the
input question
as redundant.
This relevancy
is typically
estimated
using the
attention
scores
of
textual tokens
to visual 
tokens
(referred to
as
\textit{text-to-image
attention
scores})~\cite{chen2025fastv,zhang2024sparsevlm,song2024less,xing2024pyramiddrop,shen2024longvu,liu2024multistage,zhu2024focusllava,han2024rethinking},
or 
the 
accumulative
attention scores
of
visual
tokens~\cite{he2024zipvl,ye2024fit,tu2024vlcache,jiang2024fopru}.
These 
methods
propose
classifying
visual tokens with
lower attention scores
as redundant,
as 
they are 
minimally
involved
in the cross-modal
feature interaction process.

\textbf{In summary}, 
both vision-centric
and instruction-based
strategies extensively utilize
MLLMs'
\textit{intermediate states}
to estimate
visual
redundancy.
However, 
the specific
influence
of visual tokens
with
low ViT-$[cls]$ association
or
low text-to-image attention scores
on MLLMs' output probability distribution
remains unclear.
This ambiguity
can
result in
inaccurate identification
of redundant visual tokens.

\subsection{Reducing Visual Tokens in MLLMs}

\textbf{Training-based Methods}.
Earlier
works
design 
additional 
networks
modules~\cite{instructblip,li2024tokenpacker,liu2024oryx,tong2024flashsloth,chen2024fewer,jiang2024maven}
or 
tunable
embeddings~\cite{yao2024deco,ye2024voco}
to 
compress
image patch features
into compact
embeddings,
resulting in
substantial
training cost.

\textbf{Training-free Methods}.
Recent work achieves
training-free 
visual token pruning
by
leveraging
MLLMs'
\textit{intermediate states},
discarding
visual tokens
based on
carefully
crafted
redundancy
estimation
metrics~\cite{chen2025fastv,xing2024pyramiddrop,shen2024longvu,jiang2024fopru,ye2024fit}.
Furthermore,
visual tokens can be
aggregated
into
identified
\textit{anchor tokens}
that
encapsulate
condensed 
visual
information~\cite{shang2024llavaprumerge,zhang2024sparsevlm,xu2024freepruner,yang2024enhancing,han2024rethinking},
thereby
mitigating
information loss.
However,
inaccurate
identification
of
redundant
visual tokens
can compromise
the
effectiveness
of these 
methods.
In this study, 
we propose
to 
explain
visual redundancy
by
examining
the impact of
visual tokens 
on
MLLMs'
predictions,
instead of MLLMs' 
\textit{intermediate states},
and design a
training-free
pruning strategy.
\section{Visual Redundancy Analysis}
\label{sec:analysis_redundancy}

\begin{figure*}[tb]
  \centering
  \includegraphics[width=0.9\linewidth]{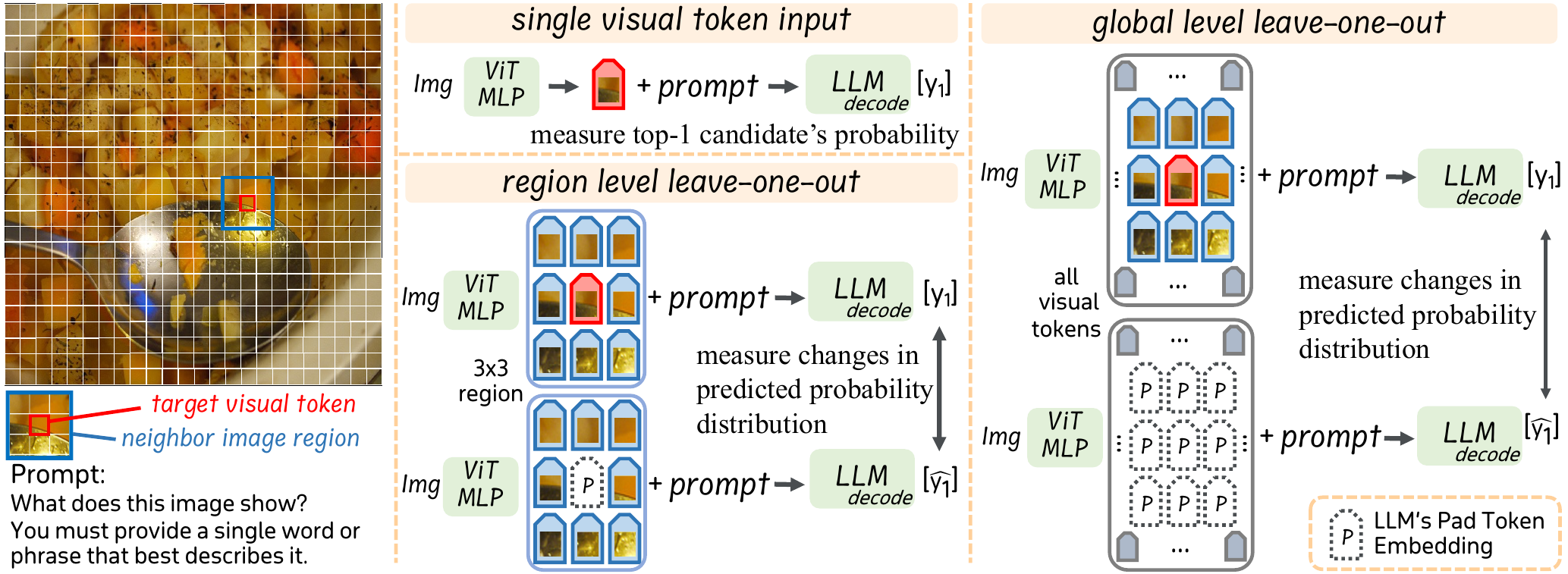}
  \caption{
    Overview of our proposed
    visual redundancy analysis pipeline.
    In the 
    \textbf{\textit{single visual token input}} experiment, 
    we 
    provide
    a single visual token
    to the LLM
    and instruct it to
    describe the visual content.
    By analyzing the predicted
    probabilities,
    we assess the significance
    of the
    information
    encoded in
    each visual token.
    Next,
    we 
    examine
    the
    influence
    of individual
    visual tokens
    on the broader
    \textit{visual context}
    (image or image region)
    by measuring
    changes in the predicted
    probability distribution
    before and after
    ablating 
    specific visual tokens.
    The \textbf{\textit{region level leave-one-out}} experiment
    evaluates
    the influence of
    a single
    visual token (highlighted in red)
    on 
    its neighboring
    image region,
    while the 
    \textbf{\textit{global level leave-one-out}} experiment
    assesses the impact of this region on the
    entire
    image.
    The results from
    these two experiments
    are combined to 
    quantify the influence
    of individual visual tokens
    on the overall
    image representation.
  }
  \label{fig:analysis_pipeline}
\end{figure*}

An
interpretable
definition
of visual redundancy
necessitates
recognizing
the direct
impact
of 
individual visual tokens
on the MLLM's
visual understanding outcome,
\ie,
the 
final probability prediction.
In this section,
we devise
novel experimental frameworks
and metrics
to 
explore
this 
often-overlooked
issue,
thereby 
providing
new insights
into
the
identification
of redundant
visual tokens
in MLLMs.

\subsection{Background and Analysis Method}

Existing
methods
estimate visual redundancy
by
extensively
utilizing
the 
scalar attention
scores
derived
from the query and key
matrices,
and infer that
a lower attention score
indicates 
a weaker correlation
between the query
and
a key feature.
However, 
these attention scores
are insufficient for
elucidating the 
exact
contribution of
visual tokens
on MLLMs'
final probability prediction,
considering
the
numerous
attention layers and heads,
the impact
of
the 
attention
value vector,
and the
feature transformation process
in auto-regressive 
LLMs,
where
the 
feature
of one token
progressively transform
into 
that
of its next token
(more details
in
Appendix 1.1).
Given these challenges, 
we shift
our 
research focus
to an 
input-output
analytical approach,
examining
variations
in model output
upon
manipulating
input visual tokens.
We 
anticipate that
this approach
will
yield
more
intuitive and interpretable
results.

Additionally,
to
rigorously
analyze 
MLLMs' 
comprehension of
visual tokens,
we propose
an approach
inspired by
human interpretation of
visual media. 
As
humans 
typically
achieve
comprehensive
image understanding
by observing
individual visual elements
and assessing their
impact on
the
overall semantic
context of the image,
we 
address
the following two
problems:

\subsubsection{Addressing
the Token-Centric Problem}
\label{sec:single_input}
In this part,
we
investigate
\textbf{what
information
does
individual
visual token
inherently possess}.
Note that
we discuss
visual information
from the viewpoint
of the LLM,
as
it
further
aggregates visual information
from
visual tokens
produced by the
vision encoder
and 
generates
textual responses.
To explore this,
we
devise a
\textit{single visual
token input experiment},
as illustrated
in~\Cref{fig:analysis_pipeline}.
We provide only one visual token
to the LLM
to eliminate the interference
from
other visual tokens
and instruct
the LLM
to describe
the visual content.
Subsequently,
we analyze
the
text decoding results
and the
predicted
probabilities
to uncover 
the LLM's
interpretation
of
the
visual
information.

To evaluate
whether
individual visual tokens
contain recognizable
information,
we assess the
magnitude of the
probability of the
$1^{st}$ ranked
textual token candidate
(denoted as 
\textit{top-1 probability}).
A
higher
\textit{top-1 probability}
indicates that
the LLM has greater confidence
in
a strong association
between
the 
$1^{st}$ ranked 
textual token
and
the input visual token.
Conversely,
if the 
\textit{top-1
probability}
is close to zero,
we 
infer
that the
visual token
does not contain
significant
visual information,
as the LLM
predicts close
confidence scores
(\ie, logits)
for
various
candidates
in the vocabulary,
indicating
high uncertainty.
Details
are in 
Appendix 1.2.

\subsubsection{Addressing the 
Context-Centric Problem}
\label{sec:leave_one_out}
We
further
investigate
\textbf{how individual
visual tokens
influence the 
overall information
of the broader
visual context
(image or image region)}
by
conducting
a
\textit{leave-one-token-out
experiment},
evaluating
the 
difference in 
the
predicted
probability distribution
before and after
the
ablation
of input visual tokens.
However,
our
preliminary 
experiments
reveal that
removing 
a single
visual token
from the image
token
sequence
results in 
numerically insignificant
changes
in the
predicted probabilities,
which
poses challenges
for
subsequent
analysis
(details in
Appendix 1.3).

To 
address this,
we devise a
\textit{cascaded 
leave-one-out experiment},
as illustrated in~\Cref{fig:analysis_pipeline}. 
First,
we conduct a
\textit{region-level
leave-one-out}
experiment
within the 3$\times$3 spatial
neighborhood of
a target visual token.
We compare the output variations
before and after
replacing the target visual token
with the LLM's 
pad token embedding
$P$.
This experiment demonstrates
the impact of
a single visual token
on the information
of
its neighboring region.
To
reveal the influence
of
this 
region
on the overall information
of the image, 
we conduct a
\textit{global-level
leave-one-out}
experiment,
inspecting
the
output variations
before and after
replacing nine
visual tokens
in 
this
region
with $P$.
By cascading the results of
these two experiments, 
we determine the influence of
individual visual tokens
on the overall information
of the image.
We employ
Jensen-Shannon Divergence
(JSD)
to assess the difference
between two
probability distributions.
The final results 
are obtained by
a weighted sum of
the JSD values from
these two
experiments.

\subsection{Discoveries}
\label{sec:discovery}

\begin{figure}[tb]
  \centering
  \includegraphics[width=0.9\linewidth]{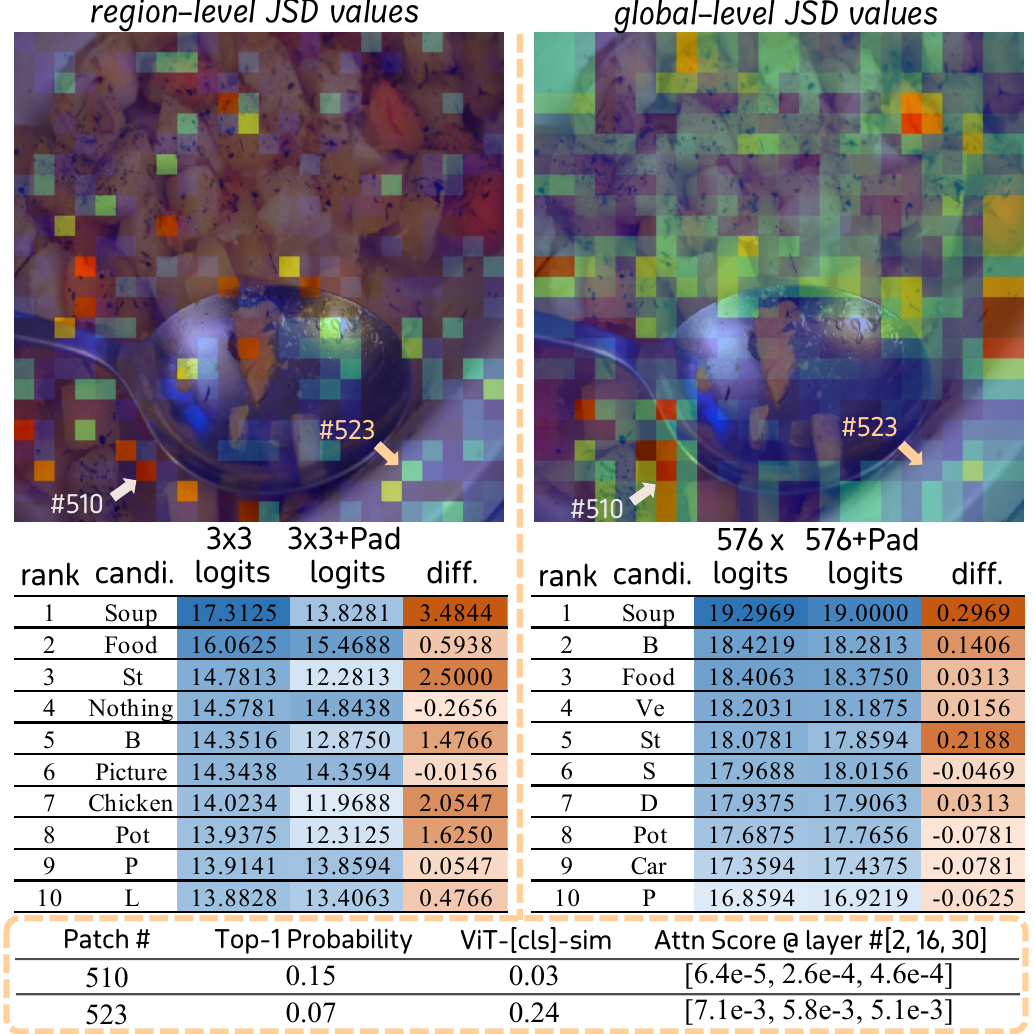}
  \caption{
    Visual tokens with
    low ViT$-[cls]$ similarity
    and text-to-image attention scores
    can more significantly impact
    LLaVA-Next's understanding of the image,
    as
    patch \#510
    has
    higher
    JSD values
    than patch \#523.
    \textit{candi.}
    and
    \textit{diff.}
    denote
    candidates
    and differences,
    respectively.
    Patch \#510
    primarily
    contributes
    the 
    semantic
    information
    \textit{Soup}
    to its
    neighboring
    region
    (+3.4844 confidence scores)
    and to the entire image
    (+0.2969 scores).
  }
  \label{fig:analysis_compare_jsd}
\vspace{-.1cm}
\end{figure}

We compare
the \textit{top-1 probability}
and the JSD results
with 
commonly
addressed
\textit{intermediate states}
in MLLMs,
including
the 
cosine
similarity
to the
$[cls]$ token
in the penultimate ViT layer
and the attention scores
of textual tokens
to visual tokens
in the 
LLM
(\ie, the \textit{text-to-image
attention scores}).
Our main findings 
are summarized as follows:

\textbf{Finding 1.
Visual tokens with
low
ViT$-[cls]$ similarity
and low
text-to-image
attention scores
may
contain
recognizable
visual information.}
For instance,
LLaVA-Next
predicts
the word 
\textit{Carrot} 
with 80\% confidence
for the image patches
depicting carrots
in the pink box
in~\Cref{fig:analysis_compare}.
However,
the
ViT$-[cls]$ similarities
and attention scores
of these patches
are only around 0.03
and in the range of
1e-5 to 1e-4,
respectively.
Conversely,
some visual tokens
with
higher
ViT$-[cls]$
similarity
and 
text-to-image attention scores
do not contain
recognizable visual information.
For instance,
LLaVA-Next
predicts irrelevant
textual responses
(\eg, \textit{Cat} and \textit{Tree})
with low confidence
(\textless10\%)
for
six 
patches
in the
uninformative white region
in the red box
in~\Cref{fig:analysis_compare},
which
have
high
ViT$-[cls]$
similarity
around 0.4
and attention scores
on the order of 1e-2.

\textbf{Finding 2.
Visual tokens with
low
ViT$-[cls]$ similarity
and
low
text-to-image
attention scores
can substantially influence
the information
of their visual context}.
For example,
patch \#510
in~\Cref{fig:analysis_compare_jsd}
significantly 
affects
the information of
its 3$\times$3 neighboring
region.
The predicted
confidence scores
(\ie, logits)
for 
specific candidates
(\eg, $Soup$ and $Chicken$)
show
notable
variation
($-2$ to $-3$ scores)
after 
patch \#510
is ablated.
This pattern
results in a
more
significant
difference
in the 
probability distribution
and a larger JSD value.
Additionally,
the neighboring region
of
patch \#510
also
notably
impacts
the
overall
image
information,
achieving
one of the
highest
JSD values
across all image regions.
However,
patch \#510's
text-to-image
attention scores
are only
at the magnitude of 1e-4,
and its ViT$-[cls]$ similarity
is merely 0.03.
In contrast,
patch \#523
has
attention scores
and ViT$-[cls]$ similarity
an order of magnitude
higher
than 
those of
patch \#510,
while ablating it
or its neighboring region
results in 
a
more
negligible
difference
in the 
model 
prediction
and
a lower JSD value.

\begin{figure}[tb]
  \centering
  \includegraphics[width=0.9\linewidth]{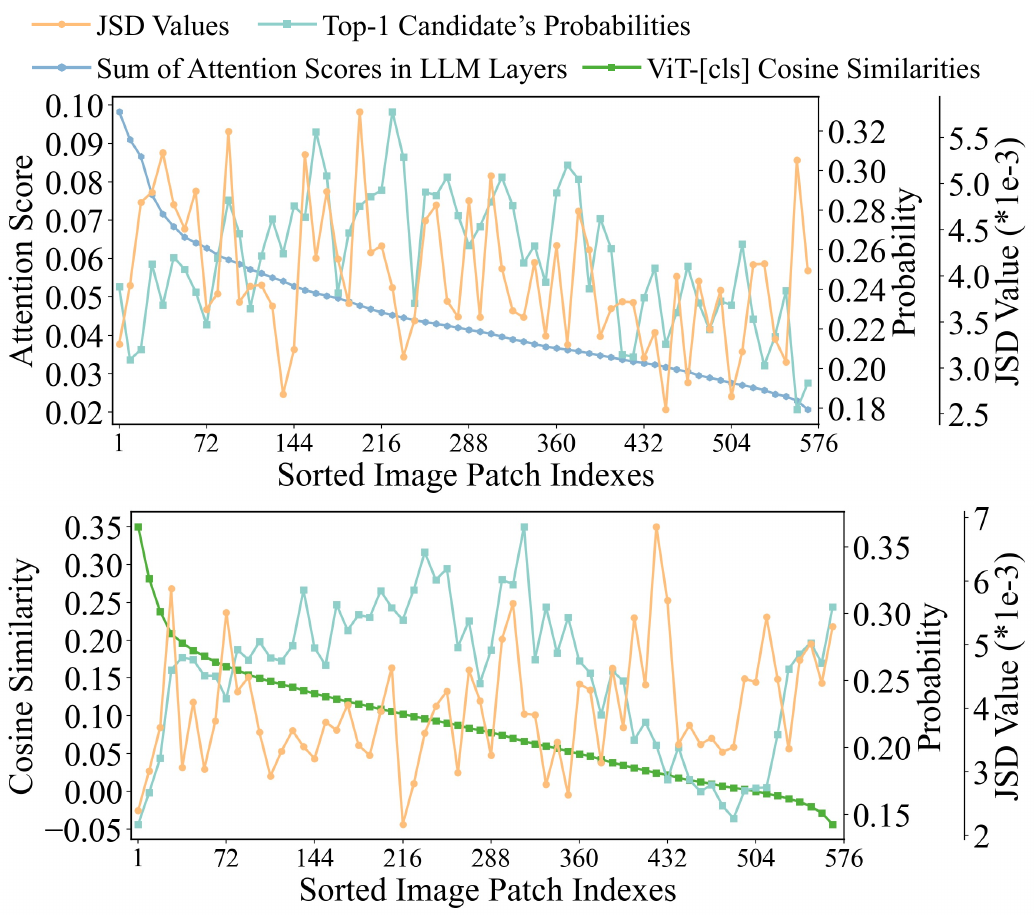}
  \caption{
    Quantitative
    results
    on 
    6,400 image patches
    sampled
    from
    the VQAv2
    validation set.
    As the text-to-image
    attention score
    and the ViT$-[cls]$ similarity
    decrease,
    the 
    \textit{top-1 probability}
    and the 
    Jenson-Shannon Divergence
    do not show a 
    declining trend;
    instead, 
    they fluctuate 
    around 0.24 and 4e-3, 
    respectively.
    The
    results are averaged
    across 100 image samples.
  }
  \label{fig:analysis_compare_quantitative}
\vspace{-.1cm}
\end{figure}

\textbf{Additional Evidences.}
To 
substantiate
the two findings,
we sample
6,400 image patches
from
the VQAv2~\cite{goyal2017making}
validation set
to conduct
\textit{single-token-input}
and 
\textit{cascaded leave-one-out}
experiments.
The results
for these image patches
are reordered
based on the
ViT$-[cls]$ similarity
or
the text-to-image 
attention score
to illustrate variation trends.
As shown
in~\Cref{fig:analysis_compare_quantitative},
when the
text-to-image
attention score
and the ViT$-[cls]$
similarity
decrease,
the 
\textit{top-1 probability}
and JSD value
do not
show
a
corresponding
decline
but rather
a fluctuating pattern.
More details and discussions are in Appendix 1.4.
Therefore, 
directly
pruning 
visual tokens
with 
low ViT$-[cls]$ similarity
or
low text-to-image attention scores
may lead to 
the loss of
visual information
and changes
in the
overall information
of the image.


\section{Method}
\label{sec:method}

\begin{figure}[tb]
  \centering
  \includegraphics[width=0.9\linewidth]{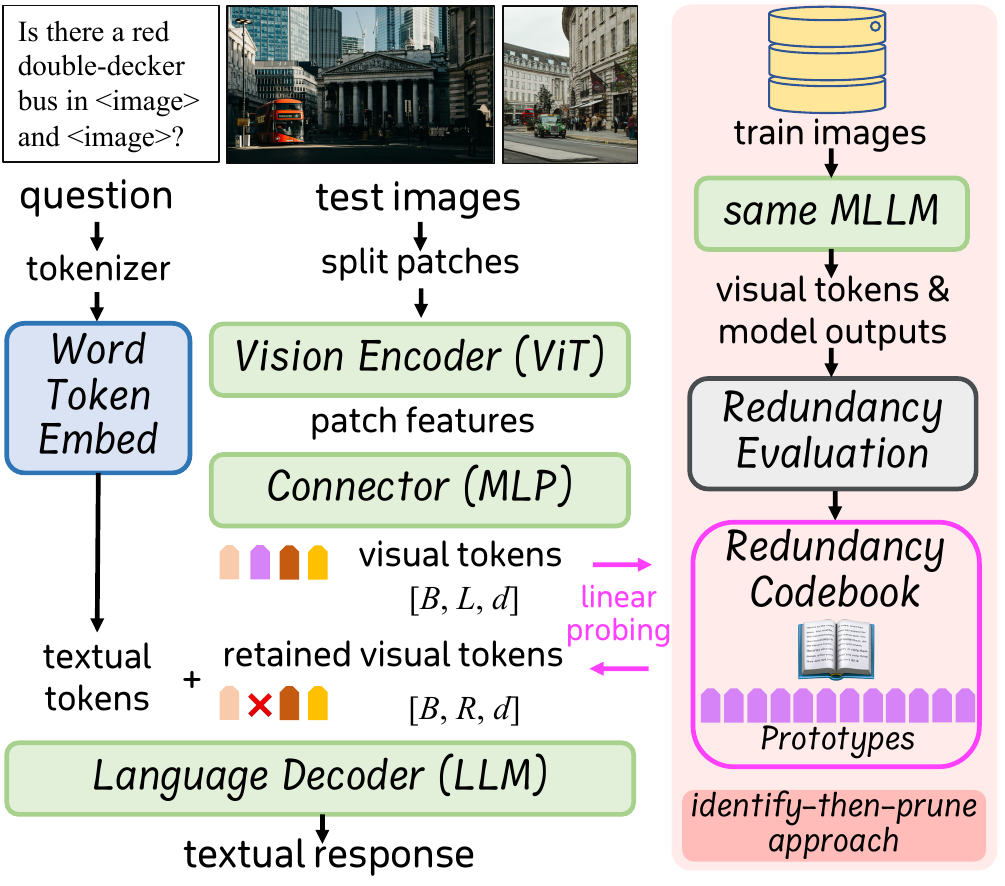}
  \caption{
    An overview of
    our 
    identify-then-probe
    approach.
    We
    identify
    \textit{redundant prototypes}
    from training images
    using 
    \textit{single-input}
    and
    \textit{cascaded leave-one-out}
    experiments,
    and store them in a
    extensible
    codebook.
    During inference,
    visual tokens
    with higher similarity
    to these prototypes
    are 
    considered
    more 
    likely to be redundant
    and are removed
    before the first layer
    of 
    the
    LLM.
    $L$ and
    $R$ are the number
    of 
    input and
    retained visual tokens,
    respectively.
  }
  \label{fig:method}
\end{figure}

Building on our 
analysis
of the direct impact
of individual visual tokens
on MLLMs'
visual understanding outcomes,
we
explore more reliable 
approach
to identify 
redundant visual tokens.
Next,
we propose an
identify-then-probe
strategy
for efficient
inference-time 
visual token pruning,
recognizing
that
\textit{single-input}
and 
\textit{leave-one-out}
experiments
entail
significant
computational overhead.
An overview
of our approach
is
depicted in~\Cref{fig:method}.
Initially,
we 
identify
\textit{redundant prototypes}
from training images
using
these two
experimental frameworks
and
store them in a
\textit{codebook}.
We then
utilize
these prototypes 
to
probe the redundancy 
of visual tokens
during inference.

\subsection{Constructing the 
Redundancy Codebook}
Based on
the impact of
individual visual tokens
on MLLMs'
visual understanding
from both
\textit{token-centric}
and 
\textit{context-centric}
perspectives,
we 
define
a potentially redundant visual token
(referred to as \textit{redundant candidate})
as one that meets
two fundamental criteria:
\textbf{(1)}
it lacks recognizable
visual information,
and
\textbf{(2)}
it does not
substantially affect
the overall information of its
associated image.
Additionally,
we observe that
certain 
\textit{redundant candidates}
from
different images
exhibit high similarity,
indicating that
these clusters
fail to contribute
substantial information
across various
\textit{visual contexts},
thus
demonstrating
potential
generalization capability.
Consequently,
we 
introduce
a 
\textit{context-independent
condition}
to
identify
\textit{redundant candidates}
with
this characteristic
as
\textit{redundant prototypes},
which
are
stored in 
an extensible
\textit{redundancy codebook}
to
facilitate
flexible
and
scalable
applications.
\subsubsection{Token-Centric
Visual Redundancy Evaluation}
The token-centric
criterion
is designed to
identify
visual tokens 
that lack
recognizable
visual information.
As discussed
in~\Cref{sec:discovery},
a low
\textit{top-1
probability}
(the
probability
of the $1^{st}$ ranked
textual
token 
candidate,
obtained
from the
\textit{single-token-input}
experiment)
indicates
the
MLLM's
inability
to 
recognize
valid
information
in individual visual tokens.
Thus,
we 
establish
a \textit{probability threshold}
$\tau_{prob}$
to 
filter out
visual tokens
with lower
\textit{top-1 probability}.

To improve
the accuracy 
in identifying
visual tokens
that lack
recognizable
visual information,
we employ t-SNE
to visualize the distribution
of visual tokens
of an image
in the high-dimensional
feature space.
We observe that
visual tokens with
very low 
\textit{top-1 probability}
frequently manifest as
discrete outliers
(as illustrated in Appendix Figure 2).
Therefore,
we use the
Density Peaks Clustering 
(DPC) 
algorithm to
find
visual tokens 
that
belong
to clusters with sizes
below a specified
\textit{outlier
threshold}
$\tau_{out}$.

\subsubsection{Context-Centric
Visual Redundancy Evaluation}
The context-centric criterion
is designed to
identify
visual tokens
with minimal
contribution
to their
\textit{visual context}.
Recall that
a low
Jensen Shannon Divergence
(JSD)
in the
\textit{cascaded 
leave-one-out}
experiment
indicates
negligible
influence of
individual
visual tokens
on MLLMs' understanding
of their
associated image
(\Cref{sec:discovery}),
we
set a 
\textit{JSD
threshold}
$\tau_{jsd}$
to 
filter out 
visual tokens with
lower JSD.
We 
then
identify
\textit{redundant candidates}
by
taking
the intersection of
visual tokens
filtered by
$\tau_{prob}$,
$\tau_{out}$,
and $\tau_{jsd}$.

\textbf{Context-independent Condition.}
After 
identifying
the 
\textit{redundant candidates}
from training images,
we further
investigate
their
capability
to
generalize
in
evaluating
the
visual redundancy
of test images.
We
analyze the distribution
of these
\textit{redundant candidates}
utilizing
t-SNE
and
observe that
some
\textit{redundant candidates}
from
different images
establish several
high-density
clusters
(as shown in Appendix Figure 3).
This phenomenon 
suggests
that,
despite differences
in the 
images,
certain
redundant candidates
share
common features.
This characteristic
indicates
potential
for
generalization.
Consequently,
we apply the
DPC algorithm
again
to 
filter out
\textit{redundant candidates}
that belong to clusters
with sizes exceeding
a specified
\textit{inlier
threshold}
$\tau_{in}$,
thereby
gathering
visual tokens
that
are unlikely to
contribute 
substantial
information
regardless of the 
\textit{visual context}
in which they 
appear.

\textbf{Summary}.
We use
the four thresholds
to filter out $N$
visual tokens
$\{\boldsymbol{v}_i\}_{i=1}^{N}$
from
training images $\boldsymbol{X}$:
\begin{equation}
    \{\boldsymbol{v}_i\}_{i=1}^{N} = CC(TC(\boldsymbol{X} | \tau_{prob},\tau_{out})|
    \tau_{jsd}, \tau_{in}),
    \label{eq:four_thres}
\end{equation}
where $\boldsymbol{v_i}\in \mathbb{R}^{d}$,
$d$ is the
feature dimension,
$TC(\cdot)$ and $CC(\cdot)$
are \textit{token-centric}
and \textit{context-centric}
redundancy evaluation methods,
respectively.
$\{\boldsymbol{v}_i\}_{i=1}^{N}$
are
the
\textit{redundant prototypes},
We
stack them 
together
to build the
\textit{redundancy codebook}
$\mathcal{C}^{N \times d}$.
We
sample
images
$\boldsymbol{X}$
from the 
\textit{Karpathy train} split
of the COCO Caption
dataset~\cite{karpathy2015deep}.

\begin{table*}[!h]
  \centering
  \scalebox{0.88}{
  \begin{tabular}{p{2.15cm}p{2.0cm}p{1.15cm}p{1.15cm}p{1.15cm}p{1.15cm}p{1.58cm}p{1.25cm}p{1.25cm}}
    \toprule
    Model & Method & POPE & MMB$^{en}$ & SEED$^{I}$ & RWQA & MME$^P$ & NoCaps & Flickr30k \\
    \midrule
    \multirow{7}*{LLaVA-1.5 \scalebox{0.8}{\color{gray}{7B}}} & w/o Split \scalebox{0.8}{\color{gray}{576$\times$}} & 85.6 & 62.9 & 65.4 & 56.1 & 1458.9 & 16.5 & 20.0 \\
    ~ & \multicolumn{8}{c}{\cellcolor{gray!12}\textit{Retain 144 visual tokens}} \\
    ~ &  PruMerge~\cite{shang2024llavaprumerge} & 75.2 & 57.7 & 55.7 & 46.8 & 1280.8 & 14.0 & 15.8 \\
    ~ &  FastV~\cite{chen2025fastv} & 79.5 &  62.2 & 61.2 & 51.2 &  1388.2 & 15.5 & 18.8 \\
    ~ &  Ours &  84.7 & 61.6 &  62.6 &  52.7 & 1369.1 &  16.4 &  20.2 \\
    ~ & \multicolumn{8}{c}{\cellcolor{gray!12}\textit{Retain 64 visual tokens}} \\
    ~ &  PruMerge~\cite{shang2024llavaprumerge} & 73.5 & 54.6 & 53.2 & 48.4 & 1228.6 & 12.9 & 14.8 \\
    ~ &  FastV~\cite{chen2025fastv} & 69.3 &  59.9 & 54.6 & 47.6 & 1150.6 & 13.4 & 15.3 \\
    ~ &  Ours &  79.9 & 57.1 &  57.3 &  48.5 &  1290.5 &  15.1 &  18.8 \\
    \cline{1-9}
    \multirow{11}*{LLaVA-Next \scalebox{0.8}{\color{gray}{8B}}} & w/o Split \scalebox{0.8}{\color{gray}{576$\times$}} & 83.9 & 72.2 & 71.4 & 56.2 & 1504.2 & 16.1 & 18.8 \\
    ~ & w Split \scalebox{0.8}{\color{gray}{2880$\times$}} & 87.8 & 72.1 & 72.7 & 59.5 & 1555.8 & 16.6 & 19.3 \\
    ~ & \multicolumn{8}{c}{\cellcolor{gray!12}\textit{w/o Split, Retain 64 visual tokens}} \\
    ~ &  Random & 76.7 \scalebox{0.6}{$\pm$0.2} & 59.2 \scalebox{0.6}{$\pm$0.7} & 62.0 \scalebox{0.6}{$\pm$0.2} & 46.7 \scalebox{0.6}{$\pm$0.9} & 1188.2 \scalebox{0.6}{$\pm$10.6} & 13.5 \scalebox{0.6}{$\pm$0.02} & 15.1 \scalebox{0.6}{$\pm$0.02} \\
    ~ &  Ours & 80.8 & 66.6 & 63.7 & 54.6 & 1224.4 & 15.1 & 17.8 \\
    ~ & \multicolumn{8}{c}{\cellcolor{gray!12}\textit{w Split, Retain 64 visual tokens per sub-image}} \\
    ~ &  Random & 81.7 \scalebox{0.6}{$\pm$0.3} & 63.3 \scalebox{0.6}{$\pm$0.4} & 65.7 \scalebox{0.6}{$\pm$0.1} & 47.9 \scalebox{0.6}{$\pm$1.1} & 1339.0 \scalebox{0.6}{$\pm$14.3} & 14.6 \scalebox{0.6}{$\pm$0.1} & 16.6 \scalebox{0.6}{$\pm$0.01} \\
    ~ &  Ours & 85.2 & 69.6 & 68.3 & 57.5 & 1343.8 & 16.1 & 18.8 \\
    ~ & \multicolumn{8}{c}{\cellcolor{gray!12}\textit{w Split, Retain 32 visual tokens per sub-image}} \\
    ~ &  Random & 77.9 \scalebox{0.6}{$\pm$0.2} & 58.4 \scalebox{0.6}{$\pm$0.4} & 62.1 \scalebox{0.6}{$\pm$0.1} & 45.5 \scalebox{0.6}{$\pm$0.2} & 1209.4 \scalebox{0.6}{$\pm$19.3} & 13.5 \scalebox{0.6}{$\pm$0.04} & 15.0 \scalebox{0.6}{$\pm$0.02} \\
    ~ &  Ours & 82.7 & 66.2 & 64.4 & 55.2 & 1254.1 & 15.2 & 17.7 \\
    \bottomrule
  \end{tabular}
  }
  \caption{Results
  on single-image VQA
  and image captioning
  benchmarks.
  The officially
  defined accuracy metric
  is reported
  for POPE, MMB-en, 
  SEED-Image, RealWorldQA 
  (RWQA)
  and MME-Perception.
  For the
  image captioning benchmarks
  NoCaps and Flickr30k,
  we report the SPICE metric.
  Our method
  outperforms
  representative
  methods
  that utilize
  MLLMs'
  \textit{intermediate states}.
  For the random baseline,
  we report the average results
  and the standard deviations
  from three separate runs.
  }
  \label{tab:single_img_res}
\end{table*}

\subsection{Pruning Visual 
Tokens using the Codebook}
In the preceding paragraphs, 
we 
have
identified
\textit{redundant prototypes}
from different images
that
exhibit
analogous
features.
Based on this
characteristic,
we infer that
visual tokens
with
higher similarity
to these
prototypes
are more likely
to be redundant,
and pruning
them
should
have
lower
impact
on MLLM's
visual understanding
outcome.
Therefore,
we utilize the
\textit{redundancy codebook}
$\mathcal{C}^{N \times d}$
to
probe
the redundancy of
$L$ input
visual tokens
$\mathcal{T}^{L \times d}$
of the test images
using the
cosine similarity:
\begin{equation}
    \mathcal{S}^{L \times N} = norm(\mathcal{T}^{L \times d}) \cdot (norm(\mathcal{C}^{N \times d}))^{T},
    \label{eq:cosine_similarity}
\end{equation}
where the $norm(\cdot)$ function
is the
L2 normalization algorithm
along the feature dimension.
We
define
the 
\textit{redundancy score}
as the
maximum cosine similarity
among the $N$ results.
Finally,
$R$
visual tokens 
with 
the lowest
\textit{redundancy scores}
are
retained
for
the LLM
($R$\textless$L$,
more details are in 
Appendix 2.2).
Different from
previous work
that employs a
huge
codebook
(\eg, $2^{17}$ 
embeddings as
in~\cite{lu2024ovis})
to 
augment the 
input
visual embeddings,
we find that
a tiny codebook
with fewer than 1,000
\textit{redundant prototypes}
generalizes well 
to test images.
Our method can be
integrated into
various
MLLMs 
without
additional
training.

\section{Experiments}
\label{sec:exp}

\subsection{Experimental Settings}

\textbf{Benchmarks and Metrics.}
We
evaluate the effectiveness
of our 
approach
on
various
vision-language
tasks,
including
single-image
Visual Question Answering
(on
POPE~\cite{li2023evaluating},
MMBench~\cite{liu2025mmbench},
SEED-Image~\cite{li2023seed},
MME~\cite{fu2024mme},
and
RealWorld-QA~\cite{rwqa}
benchmarks),
image captioning
(NoCaps-val~\cite{agrawal2019nocaps}
and Flickr30k-test~\cite{flickr30k}),
and
multi-image and video
comprehension
(Mantis-test~\cite{jiang2024mantis},
MuirBench~\cite{wang2024muirbench},
and MVBench~\cite{li2024mvbench}).
We adhere to
the officially
defined
metrics
(Exact Match Accuracy)
for
VQA,
and utilize the
SPICE~\cite{anderson2016spice}
metric for image captioning,
which emphasizes semantic correspondence.

\textbf{Implementation Details.}
We implement our method
on three MLLMs:
LLaVA-1.5~\cite{liu2023llava,liu2024improved},
LLaVA-Next~\cite{liu2024llavanext,li2024llavanext-strong},
and LLaVA-OneVision~\cite{li2024llavaov}.
For each model,
we
construct 
a distinct
codebook,
as
model predictions
are necessary 
to evaluate
the
contribution
of visual tokens.
We 
set a 
threshold
to
remove visual tokens
with 
the highest
\textit{redundancy score}.
We employ the
greedy decoding
method
for reproducible results.
Detailed settings
are in Appendix 3.1.

\subsection{Experimental Results}
We compare 
the performance
of
our 
method
with
two
representative
approaches
that
leverage
MLLMs'
\textit{intermediate states}:
the
vision-centric
method PruMerge~\cite{shang2024llavaprumerge},
which prunes
visual tokens with lower
association with the ViT$-[cls]$ token,
and the
instruction-based
method FastV~\cite{chen2025fastv},
which
leverages the
attention scores
of the last textual token
to visual tokens
within the
LLM.
For a fair comparison, 
we maintain a
training-free 
setting
and adhere to
the same
visual token quantity budgets.

\subsubsection{Single-Image Comprehension}
Results
on single-image VQA
and captioning tasks
are presented
in~\Cref{tab:single_img_res}.
Notably, 
for the LLaVA-1.5 model,
our method
preserves
90\% of
peak
performance
(\ie, with 576 
input visual tokens)
on 
average
across five VQA
benchmarks
while retaining
only 11\% of
visual tokens. 
In contrast, 
both the 
vision-centric
and
instruction-based
strategies
achieve
approximately 85\%.
When retaining 25\% 
of
visual tokens, 
our method
maintains or
slightly exceeds
the performance ceiling
on two 
image captioning benchmarks,
significantly outperforming
the vision-centric strategy
(82\% performance) 
and the instruction-based strategy
(94\% performance).
Under both the
sub-image splitting
and
non-splitting settings
of the LLaVA-Next model, 
our method preserves
95\% and 91\% 
performance,
respectively, 
while retaining only 11\%
of 
visual tokens. 
In contrast, 
the random pruning baseline
achieves 87\% and 84\%. 
Additionally,
our method maintains 90\%
performance
for the LLaVA-Next model
under a very low
retention rate of
visual tokens (5.5\%).
These results
demonstrate 
that
assessing visual redundancy
based on MLLMs' predictions
is superior to
utilizing 
MLLMs'
intermediate states.
Qualitative
results
in
Appendix Figures 5 to 9
show
that
our method
allocates
the
limited visual token budget
to critical
visual cues
in
both
natural photographs
and text-rich images.

\subsubsection{Multi-Image and Video Comprehension}

\begin{table}[tb]
  \centering
  \scalebox{0.85}{
  \begin{tabular}{p{1.3cm}p{2.0cm}p{2.0cm}p{2.0cm}}
    \toprule
    Method & Mantis-test & MUIRBench & MVBench \\
    \midrule
    ~ & \multicolumn{2}{c}{\cellcolor{gray!12}\textit{729 per image}} & \cellcolor{gray!24}\textit{196 / img} \\
    w/o Split & 59.0 \scalebox{0.7}{\color{gray}{(1814$\times$})} & 42.7 \scalebox{0.7}{\color{gray}{(3158$\times$})} & 58.7 \scalebox{0.7}{\color{gray}{(3136$\times$})} \\
    ~ & \multicolumn{2}{c}{\cellcolor{gray!12}\textit{Retain 144 per image}} & \cellcolor{gray!24}\textit{16 / img} \\
    Random & 61.4 \scalebox{0.7}{$\pm$0.6} \scalebox{0.7}{\color{gray}{(358$\times$})} & 45.2 \scalebox{0.7}{$\pm$0.1} \scalebox{0.7}{\color{gray}{(624$\times$})} & 53.2 \scalebox{0.7}{$\pm$0.3} \scalebox{0.7}{\color{gray}{(256$\times$})} \\
    Ours & 63.6 \scalebox{0.7}{\color{gray}{(351$\times$})} & 48.1 \scalebox{0.7}{\color{gray}{(626$\times$})} & 55.0 \scalebox{0.7}{\color{gray}{(256$\times$})} \\
    \bottomrule
  \end{tabular}
  }
  \caption{LLaVA-OneVision-7B
  results on 
  multi-image
  and video comprehension
  benchmarks.
  Our proposed
  method maintains over
  90\% of
  peak performance 
  and 
  achieves
  a
  10\% performance gain
  by pruning 80\% to 90\% of
  input visual tokens.
  }
  \label{tab:multi_img_res}
\end{table}

Results 
on multi-image
and video comprehension
tasks are 
presented
in~\Cref{tab:multi_img_res}.
On
Mantis-test and MuirBench,
the performance of
LLaVA-OneVision
improves by 5\%
after randomly removing
80\% of visual tokens,
while our method
achieves a higher enhancement
of 10\%.
This suggests that
an excessive number of visual tokens
may impede the model's
ability to comprehend
image-text-interleaved contexts.
In the MVBench 
video understanding benchmark, 
our approach maintains
94\% performance
even with an extreme
visual token removal
rate
of 92\%,
significantly surpassing
the random baseline.
These 
results
demonstrate that our method
can effectively transfer
from single-image 
to
multi-image
and
video comprehension tasks.

\subsection{Efficiency Analysis}
During inference,
the
primary
computational overhead
introduced
by our
method
is
the calculation of
the similarity matrix
$\mathcal{S}^{L \times N}$,
which 
incurs a
marginal
cost of
$L\times N\times(2d-1)$
floating-point operations
(FLOPs).
The codebook
requires
approximately
0.5 GB
of GPU memory.

\subsection{Ablation Study}

\begin{table}[tb]
  \centering
  \scalebox{0.85}{
  \begin{tabular}{p{0.6cm}p{0.6cm}p{0.6cm}p{0.5cm}p{0.9cm}cc}
    \toprule
    $\tau_{prob}$ & $\tau_{jsd}$ & $\tau_{out}$ & $\tau_{in}$ & \# Img. & $N$ & Avg. Perf. \\
    \midrule
    \rowcolor{green!5}
    \checkmark & \checkmark & \checkmark & \checkmark & 100\% & 969 & 91.3\% \\
    - & \checkmark & \checkmark & \checkmark & 100\% & 5,086 & 84.9\% \\
    \checkmark & - & \checkmark & \checkmark & 100\% & 1,474 & 90.6\% \\
    \checkmark & \checkmark & - & \checkmark & 100\% & 2,884 & 91.2\% \\
    \checkmark & \checkmark & \checkmark & - & 100\% & 1,151 & 90.1\% \\
    \checkmark & \checkmark & \checkmark & \checkmark & 20\% & 185 & 88.0\% \\
    \multicolumn{6}{c}{\cellcolor{gray!12}\textit{random baseline}} & \cellcolor{gray!12} 84.5\% \\
    \bottomrule
  \end{tabular}
  }
  \caption{Ablation 
  studies
  on five single-image
  VQA benchmarks
  of LLaVA-Next.
  Each component in
  our proposed method
  contributes 
  positively to the 
  average
  performance
  (Avg. Perf.).
  ``\# Img.'' denotes
  the percentage of 
  sampled images
  used to identify
  \textit{redundant prototypes}.
  $N$ is the number of
  \textit{redundant prototypes}.
  }
  \label{tab:ablation}
\end{table}

We assess the effectiveness
of each component
($\tau_{prob}$,
$\tau_{jsd}$,
$\tau_{out}$,
and
$\tau_{in}$)
in
our proposed method
by 
individually
ablating
them
and evaluating the
average performance
on five single-image VQA
benchmarks.
\Cref{tab:ablation}
demonstrates that
each component contributes positively
to the overall performance.
Notably, 
the removal of 
$\tau_{prob}$
leads to a significant
performance drop
for LLaVA-Next
(decreasing
from 91.3\%
peak
performance to 
84.9\%,
approaching
the random baseline).
In contrast, 
the performance degradation
caused by the removal of
other components
is relatively moderate. 
Additionally, 
reducing the number
of sampled training images
decreases
the number of
\textit{redundant prototypes}
from 969 to 185, 
accompanied by a 
3.3\% performance decline. 
Consequently, 
we opt to use
the 969 identified
\textit{redundant prototypes}
for LLaVA-Next.




\section{Conclusion}
We explore 
interpretable
definition
of visual redundancy
in MLLMs,
focusing on
the 
influence of 
individual visual tokens
on MLLMs'
visual understanding outcome,
which is a
often-overlooked
issue.
To 
intuitively
and
comprehensively
investigate this issue,
we
develop
input-to-output
analytical approaches
from both
\textit{token-centric}
and
\textit{context-centric}
perspectives.
We
reveal that
visual tokens
with low
ViT$-[cls]$ similarity
and
low
\textit{text-to-image
attention scores}
can contain
recognizable
visual 
information
and
substantially
influence 
their 
\textit{visual context}.
Building 
on
these findings,
we propose 
a novel
method
to
identify
redundant visual tokens
by
combining
the
\textit{token-centric}
and 
\textit{context-centric}
criteria,
along with
a
\textit{context-independent
condition}.
Utilizing
this
redundancy evaluation method,
we design an efficient and scalable
identify-then-probe approach
for training-free
visual token pruning.
On single-image,
multi-image
and video
comprehension
benchmarks,
our 
method
achieves 
90\% to 110\%
performance
while pruning
80\% to
90\% of
visual tokens,
surpassing
existing
methods
that rely on MLLMs'
\textit{intermediate states}.

{
    \small
    \bibliographystyle{ieeenat_fullname}
    \bibliography{main}
}
\clearpage
\setcounter{page}{1}
\maketitlesupplementary

\section{Visual Redundancy Analysis Details}
We conduct
token-centric
and
context-centric
experiments
on
three MLLMs:
LLaVA-1.5, 
LLaVA-Next, 
and LLaVA-OneVision,
subsequently constructing
separate redundancy codebooks
for each model.
For LLaVA-Next and LLaVA-OneVision,
the evaluation
is conducted
without splitting input images
into sub-images, 
retaining only the
base image features.

\subsection{Background on
the Attention Mechanism}
\label{appendix_sec:attn_bkgd}
In a
multi-head
self-attention layer,
the output
of
the
attention operator
is:
\begin{align}
    & \text{Attention}_{(l,h)}(\boldsymbol{Q}_{(l,h)}, \boldsymbol{K}_{(l,h)}, \boldsymbol{V}_{(l,h)}) \notag \\
    &= \mathrm{softmax}\left(\frac{\boldsymbol{Q}_{(l,h)} (\boldsymbol{K}_{(l,h)})^T}{\sqrt{d_k}}\right) \boldsymbol{V}_{(l,h)},
\end{align}
where 
$l$ is the attention layer index,
$h$ is the attention head index,
$d_k$ is the 
head
dimension.
Considering
the
\textit{text-to-image}
self-attention computation
for the first 
decoded text token
at layer $l$ and head $h$,
the query
$\boldsymbol{q}_{(l,h)}^{1\times d_k}$ 
is linearly projected
from the
hidden state of
the text token,
while
the Keys $\boldsymbol{K}_{(l,h)}^{L\times d_k}$
and
the Values $\boldsymbol{V}_{(l,h)}^{L\times d}$
are derived from
the visual tokens.
The $j^{th}$ visual token
contributes
a feature
$\Delta \boldsymbol{h}$
to the hidden state of the query:
\begin{align}
\Delta \boldsymbol{h}^{1\times d}_{(l,h)} = \mathrm{softmax}\left(\frac{\boldsymbol{q}^{1\times d_k}_{(l,h)}(\boldsymbol{K}_{(l,h)}^{L\times d_k})^T}{\sqrt{d_k}}\right)[j] \cdot \boldsymbol{V}_{(l,h)}^{L\times d}[j, :],
\label{eq:text2img_attn}
\end{align}
where $L$ is the number
of input visual tokens.
The 
$j^{th}$ result
produced 
by the $\mathrm{softmax}$ operator
is referred to as the 
text-to-image
attention score
of the $j^{th}$ visual token
at layer $l$ and head $h$.
According to~\Cref{eq:text2img_attn},
the new feature vector
$\Delta \boldsymbol{h}$
is obtained 
by scaling the value vector
$V_{(l,h)}^{L\times d}[j, :]$,
which
has a
feature dimension of $d$
(\eg, $d$=4,096 for LLaVA-1.5).
The high dimensionality
poses challenges
to analyze
the impact of the
$j^{th}$ visual token
on the decoded text token.
Moreover,
the large number of
attention heads
(\eg, $h$=32 in LLaVA-1.5)
further
exacerbates this difficulty.

In addition to
the challenges posed by
the high dimensionality 
of hidden states
and
the large number of
attention heads
and layers,
we highlight that
as $l$ increases,
the hidden state
of one token
$\boldsymbol{H}_{(l)}^{L\times d}[i, :]$
progressively
transforms into that
of the next token
$\boldsymbol{H}_{(l=0)}^{L\times d}[i+1, :]$
in an auto-regressive language model.
Consequently,
it becomes difficult
to determine
whether the 
the attention scores
in the middle layers
of the LLM
should be attributed to
the $i^{th}$
or 
$(i+1)^{th}$
visual token.
To avoid
this ambiguity,
we propose
two novel
input-to-output 
experimental frameworks
to evaluate the
impact of
input visual tokens
on MLLMs' textual output.

\subsection{Single Visual Token Input Experiment Details}
\label{appendix_sec:single_input}
The LLaVA model family
employs
auto-regressive language 
models~\cite{zheng2023judging,llama3modelcard}
as text decoders,
selecting each token
sequentially 
based on the predicted
probability of each token candidate $x_i$
from the vocabulary $\mathcal V$:
\begin{equation}
p(x_i|\boldsymbol{v},\boldsymbol{x},\boldsymbol{y}_{<t}) = \frac{\exp{(\boldsymbol{h}_t \cdot E_c(x_i)})}{\sum_{x' \in \mathcal V} \exp{(\boldsymbol{h}_t \cdot E_c(x')})},
  \label{eq:softmax}
\end{equation}
where $\boldsymbol{v}$ is the visual input,
$\boldsymbol{x}$ and $\boldsymbol{y}_{<t}$ are the prompt and past generated tokens, respectively.
$E_c(x_i)$ is the token embedding of 
candidate $x_i$ in the language model head
(the final linear layer).
$\boldsymbol{h}_t$ is the
hidden state predicted by 
the last transformer
block
at decoding step
$t$.
$(\cdot)$ is the inner product operator.
$(\boldsymbol{h}_t \cdot E_c(x_i))$ 
is the \textit{confidence score}
(\ie \textit{logit}),
which 
manifests
MLLMs'
level of confidence
for predicting
token candidate $x_i$
conditioned on
$\boldsymbol{v}$,
$\boldsymbol{x}$,
and $\boldsymbol{y}_{<t}$.
A 
high
confidence score
results in a
higher probability
after
the $\mathrm{softmax}$ operation
(\Cref{eq:softmax}).
Token candidates
with higher probability
are more likely to be selected
during decoding.
Conversely,
if the MLLM
assigns similar
confidence scores
to multiple
token candidates,
the
$\mathrm{softmax}$ output
distributes
lower probabilities
among them,
indicating
uncertainty
in
selecting the
appropriate token
for the current decoding step.

In summary,
the
probability
of the $1^{st}$ ranked
token candidate
reflects the
MLLMs'
confidence
in
predicting
this token
given
$\boldsymbol{v}$,
$\boldsymbol{x}$,
and $\boldsymbol{y}_{<t}$.
In the single-input experiment,
$\boldsymbol{x}$
consists only of the
task description
and response format requirements.
Thus,
we use the
predicted probability
for the $1^{st}$ ranked candidate,
\ie, 
the
\textit{top-1 probability}
$p_1$,
to assess whether
the visual input
$\boldsymbol{v}$
contain recognizable
visual
information,
\begin{equation}
p_1 = \mathrm{max}(\{p(x_i|\boldsymbol{v}_{single},\boldsymbol{x},\boldsymbol{y}_{<t})|x_i \in \mathcal V\}),
\label{eq:top1_prob}
\end{equation}
where the decoding step $t=1$
(as
we explicitly instruct
the language model
to describe the 
visual content
using a single word or phrase).
If the first generated token
is an article
(\eg, \textit{The}, \textit{A}
and \textit{An}),
then
the second decoded token
is considered
($t=2$).
In the
single-input experiment
for LLaVA-1.5,
we retain only one visual token
$\boldsymbol{v}_{single}^{1\times d}$
and instruct the LLM
to describe the visual content.
However,
for LLaVA-Next and LLaVA-OneVision,
we observe that 
they often
refuse to generate responses
when
provided with
only
a single visual token.
To address this issue,
we repeat 
$\boldsymbol{v}_{single}^{1\times d}$
24 times
($\sqrt{576}$)
for LLaVA-Next
and 
27 times
$\sqrt{729}$ 
for LLaVA-OneVision~\footnote{LLaVA-Next's
CLIP-ViT vision encoder processes 576 tokens,
while LLaVA-OneVision's
SigLIP-ViT processes 729 tokens.}.
We then append a special
\textit{image newline} token
after the repeated visual tokens,
following 
the official setting.
This operation
constructs a
``\textit{synthesized image line}''
that only contains the
visual information
of the original single visual token,
leading to more reliable results.


\subsection{Leave-One-Token-Out Experiment Details}
\label{appendix_sec:leave_one_out}
\begin{figure}[tb]
  \centering
  \includegraphics[width=0.9\linewidth]{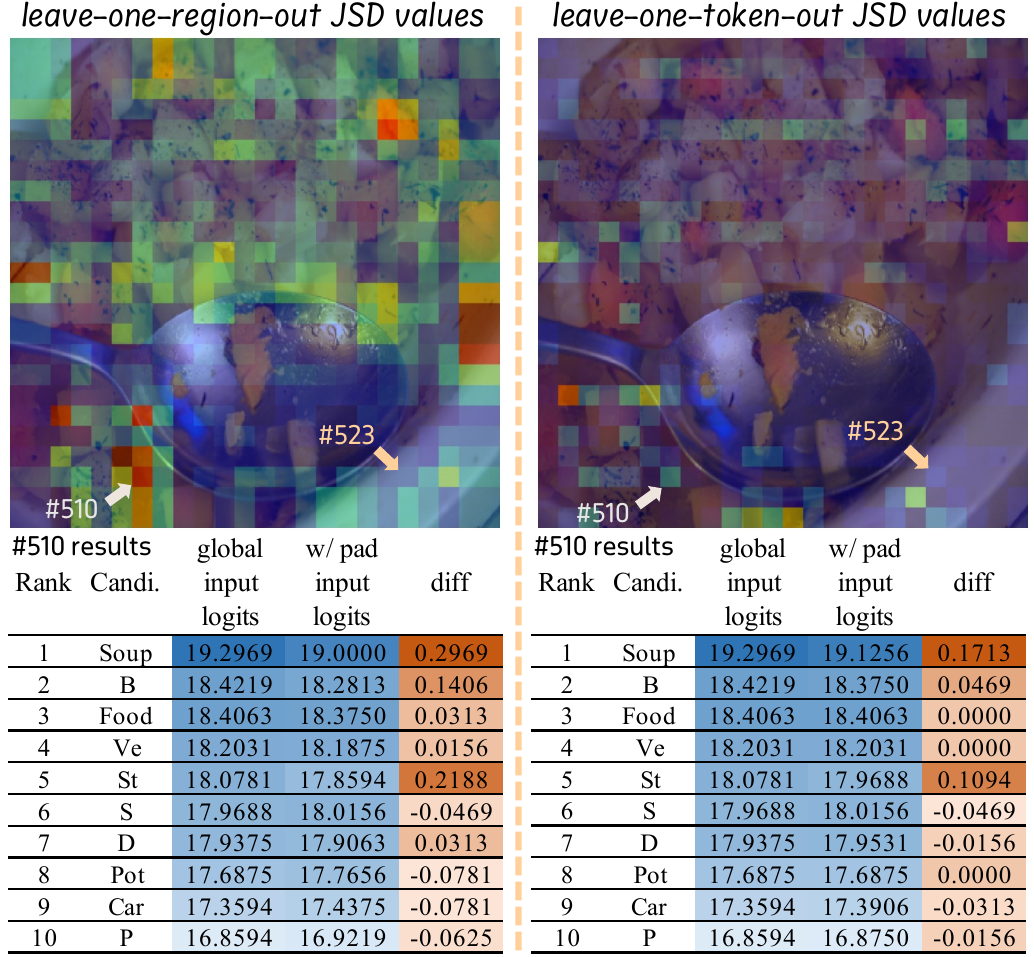}
  \caption{
    Comparison
    between
    the 
    \textit{leave-one-region-out}
    and
    \textit{leave-one-token-out}
    experiments.
    The
    \textit{leave-one-token-out}
    experiment
    results in
    numerically insignificant
    results,
    which brings challenge
    for further analysis.
  }
  \label{fig:analysis_compare_jsd_cascaded_naive}
\end{figure}

We employ
Jensen-Shannon Divergence
(JSD) 
to quantify the difference
in predicted probability distributions
before and after
ablating
individual input visual tokens,
\begin{equation}
    \mathrm{JSD}(M \parallel N) = \frac{1}{2} \left( D_{KL}(M \parallel Q) + D_{KL}(N \parallel Q) \right)
    \label{eq:jsd1},
\end{equation}
\begin{equation}
    Q = \frac{1}{2}(M + N),
    \label{eq:jsd2}
\end{equation}
\begin{equation}
    M = \mathrm{softmax}(\{\mathit{logits}(x_i|\boldsymbol{v}^{src}, \boldsymbol{x},\boldsymbol{y}_{<t})|x_i \in \mathcal V_{head}^{m}\}),
    \label{eq:jsd3}
\end{equation}
\begin{equation}
    N = \mathrm{softmax}(\{\mathit{logits}(x_i|\boldsymbol{v}^{P}, \boldsymbol{x},\boldsymbol{y}_{<t})|x_i \in \mathcal V_{head}^{m} \}),
    \label{eq:jsd4}
\end{equation}
where $D_{KL}(\cdot)$ is the KL divergence.
$\mathcal V_{head}^{m}$
is a head vocabulary
consisting
of $m$ top-ranked candidates.
$\boldsymbol{v}^{src}$
and
$\boldsymbol{v}^{P}$
are the input visual token sequences
before and after
replacing
certain
visual tokens
with the pad token embedding $P$.
The decoding step
$t=$1
(articles are also skipped here).
To assess the impact
of an individual visual token
on the predicted
probability distribution,
the JSD values
from the
\textit{region-level}
and
\textit{global-level
leave-one-out} experiments
are linearly combined:
\begin{equation}
    \mathrm{JSD}^{final} = k^{region} \mathrm{JSD}^{region} + k^{global} \mathrm{JSD}^{global},
    \label{eq:jsd_weight_sum}
\end{equation}
where $k^{region}$ and $k^{global}$
are hyper-parameters
and are set at 1 and 16
in all experiments,
respectively.
The final results $\mathrm{JSD}^{final}$
is used for comparison
with the \textit{JSD threshold}
$\tau_{jsd}$.

In preliminary experiments,
we directly replace
individual visual tokens
with the
pad token embedding
$P$
and
evaluate
the resulting changes
in the model's output.
\Cref{fig:analysis_compare_jsd_cascaded_naive}
demonstrates that
this approach
leads to
minimal numerical changes
in the model's output,
with the JSD values for
most image patches
ranging from 1e-6 to 1e-5.
For patch \#510,
the predicted
confidence scores of 
many token candidates
exhibit changes
close to zero. 
We are
concerned that
these small discrepancies
might
have
computational errors
that could interfere with the results. 
Additionally, 
in the direct
leave-one-token-out experiment, 
the JSD values
for different image patches
exhibit negligible variation
(\eg, the JSD values
for patches \#510
and \#523
differ by approximately 1e-5).
These small differences
pose challenges
for determining an appropriate
\textit{JSD threshold}
$\tau_{jsd}$.
As a result,
we propose a 
\textit{cascaded leave-one-out}
experimental scheme.


\subsection{Experiment Details on
VQAv2 Validation Set}
\label{appendix_sec:vqav2_exp}
We randomly select
100 samples from the
VQAv2 validation set
For each image,
we uniformly sample
64 patches
from the 2D image grid,
resulting in a total of
6400 patch samples.
These samples
are used
to conduct the
\textit{single-input}
and
\textit{leave-one-token-out}
experiments.
Four metrics are obtained from these experiments:
\begin{itemize}
\item The
\textit{top-1 probability}
for each patch
is calculated using~\Cref{eq:top1_prob}.
\item The
JSD result
is obtained by~\Cref{eq:jsd_weight_sum}.
\item The 
text-to-image attention scores
are first computed using~\Cref{eq:text2img_attn}.
We then
average the results
across all heads
and sum the attention scores
from a shallow layer
($l$=1),
a medium layer
($l$=16),
and a deep layer
($l$=30)
of the LLM.
\item The
ViT$-[cls]$ similarity
is computed
using the cosine similarity
between the
image patch token
and
the
$[cls]$ token
in the
penultimate ViT layer
(visual tokens 
produced by this layer
are subsequently sent to
LLaVA's cross-modal connector).
\end{itemize}
After computing the
four metrics
for each image patch, 
we
reorder the patch indices
within each image
based on either the
ViT$-[cls]$ similarity
or the
text-to-image attention score. 
We then 
aggregate
the results
according to these
reordered patch indices
and calculate
the average
across the 100 samples.

\textbf{Further Discussion.}
The results presented 
in Figure 4 in the main paper
indicate that 
certain patch tokens,
which exhibit the highest similarity
to the ViT’s $[cls]$ token,
often correspond to very low
\textit{top-1 probability} values.
This suggests that when these visual tokens
are independently fed into the MLLM,
the model fails to recognize 
valid visual information from them.
According to the study in~\cite{darcet2024vision}, 
this phenomenon may stem from a
``register'' effect in the ViT model,
where it utilizes background patches,
which carry little information,
as registers to store visual information
from other patches.
Removing these visual tokens 
helps mitigate the
\textit{high-norm artifacts}
in the image representation.

\section{Method Details}
\subsection{Constructing the Redundancy Codebook}
\label{appendix_sec:detail_for_codebook}
\begin{algorithm}[tb]
\caption{Pseudo code for constructing
the redundancy codebook}
\begin{algorithmic}[1]
\STATE \textcolor{darkgreen}{\textit{\# initialize the codebook}}
\STATE $codebook\_candidates$ = []
\STATE $codebook$ = []
\STATE \textcolor{darkgreen}{\textit{\# get visual tokens}}
\STATE $image\_feats$ = vision\_tower($image$)
\STATE $\{\boldsymbol{v}_n\}_{n=1}^{L}$ = mm\_projector($image\_feats$) 
\FOR {$n=0; n<L; n++$}
\STATE \textcolor{darkgreen}{\textit{\# perform the \textit{token-centric redundancy evaluation} utilizing the \textit{single visual token input} experiment}}
\STATE $p_1^n$, $c_{img}^n$ = $TC(\boldsymbol{v}_n,DPC(\{\boldsymbol{v}_n\}_{n=1}^{L}, n))$
\STATE \textcolor{darkgreen}{\textit{\# perform the \textit{context-centric redundancy evaluation} utilizing the \textit{cascaded leave-one-token-out} experiment}}
\STATE $JSD_n^{final}$ = $CC(\boldsymbol{v}^{src}_n, \boldsymbol{v}^{P}_n, n)$
\STATE \textcolor{darkgreen}{\textit{\# regard this visual token as a \textit{redundant candidate}}}
\STATE $low\_info\_flag$ = ($p_1^n<\tau_{prob}$) \textbf{and} ($c_{img}^n < \tau_{out}$) \textbf{and} ($JSD_n^{final} < \tau_{jsd}$) 
\IF {$low\_info\_flag$}
\STATE $codebook\_candidates$.append($\boldsymbol{v}_n$)
\ENDIF
\ENDFOR
\FOR{i \textbf{in} range(len($codebook\_candidates$))}
\STATE $c^i_{candi}$ = DPC($codebook\_candidates$, i)
\IF{$c^i_{candi}> \tau_{in}$}
\STATE \textcolor{darkgreen}{\textit{\# append the visual token to the codebook}}
\STATE $codebook$.append($codebook\_candidates[i]$)
\ENDIF
\ENDFOR
\STATE save\_to\_disk($codebook$)
\end{algorithmic}
\label{alg:build_codebook}
\end{algorithm}

\Cref{alg:build_codebook}
outlines the
procedure for constructing
the
\textit{redundancy codebook}.
Given
the visual tokens
$\{\boldsymbol{v}_n\}_{n=1}^{L}$
obtained from the vision encoder
($vision\_tower$)
and the cross-modal connector
($mm\_projector$),
we employ the
\textit{single-input}
experiment
(as described in~\Cref{eq:top1_prob})
to compute the
\textit{top-1 probability}
$p_1^n$
and
apply the
Density Peaks Clustering
(DPC)
algorithm
to determine
the size of the cluster
($c_{img}^n$)
containing the
visual token
$\boldsymbol{v}_n$.
Next, the
\textit{cascaded 
leave-one-token-out}
experiment
is performed
to compute the final
JSD value
$JSD^{final}_n$.
If $p_1^n$ 
is below the 
\textit{probability threshold} $\tau_{prob}$,
we conclude that 
$\boldsymbol{v}_n$
does not contain recognizable
visual information.
If $c_{img}^n$
is below the 
\textit{outlier threshold}
$\tau_{out}$,
$\boldsymbol{v}_n$
is classified as an outlier
in the feature embedding space.
Additionally,
if  $JSD_n^{final}$
is below the
\textit{JSD threshold} $\tau_{jsd}$,
we assert that
$\boldsymbol{v}_n$
has negligible
impact on the 
overall information of 
its associated image.
If all three conditions are satisfied,
we classify $\boldsymbol{v}_n$
as a \textit{redundant candidate}
(lines 13 to 16 in~\Cref{alg:build_codebook}).
After identifying
all \textit{redundant candidates},
we apply
the DPC algorithm again
to detect groups of similar images
that are unlikely to 
contribute substantial
visual information,
regardless of the image
in which they appear.
Specifically,
we retain only the
redundant candidates
that belong to
clusters with a size larger than 
the \textit{inlier threshold} $\tau_{in}$.
Finally,
we stack the
identified \textit{redundant prototypes}
and save them to disk.

\begin{figure}[tb]
  \centering
  \includegraphics[width=0.9\linewidth]{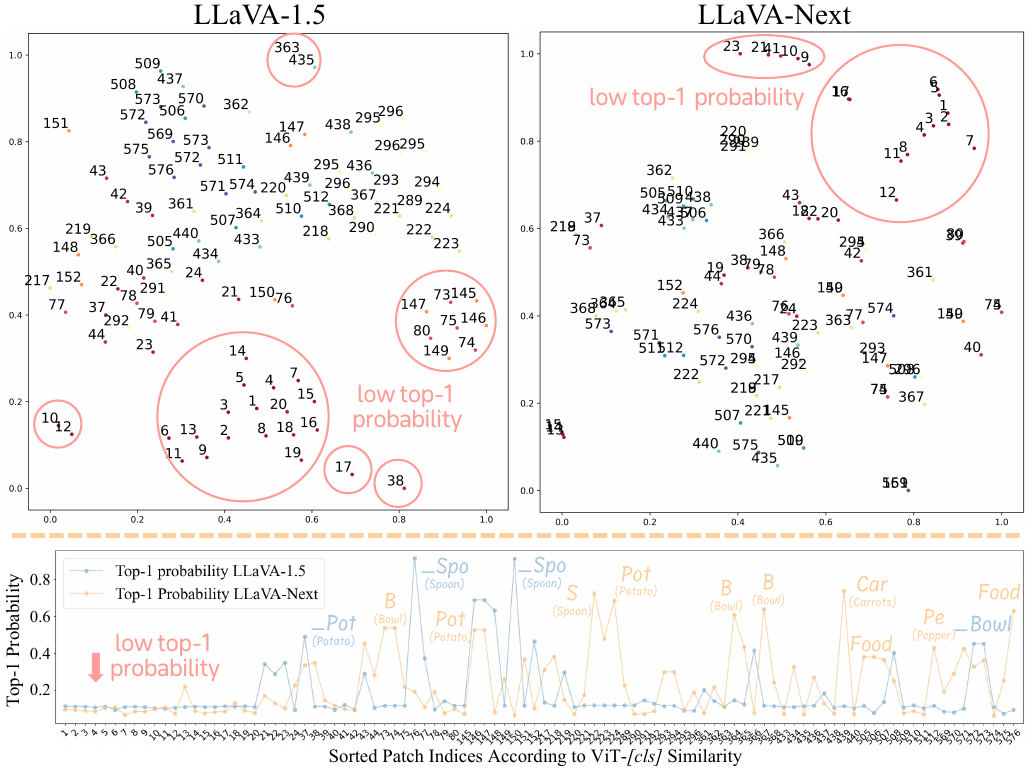}
  \caption{
    t-SNE visualization
    of visual token
    distribution
    in feature space
    for a single image.
    Visual tokens with low
    \textit{top-1 probability}
    often appears
    as discrete outliers
    (indicated by the pink circles).
    Additionally,
    image patch tokens
    exhibiting high similarity
    to the ViT$-[cls]$ token
    generally have lower
    \textit{top-1 probability},
    suggesting a lack of distinguishable
    visual information. 
    We also present the 
    greedy decoding results
    for image patch tokens
    with higher
    \textit{top-1 probability}.
  }
  \label{fig:tsne_single_img}
\end{figure}

We use
t-SNE for
dimensionality reduction
to analyze
the distribution
of visual tokens
from a single image
in high-dimensional feature space.
\Cref{fig:tsne_single_img}
shows
that visual
tokens with low
\textit{top-1 probability}
($\leq$0.1)
often appear as
outliers,
while those with
higher
\textit{top-1 probability}
tend to form larger clusters.
This observation motivates
us to apply
clustering algorithms
to identify outlier visual tokens
within an image.
This step
prevents
redundant candidates
that are similar to
visual tokens
containing recognizable visual information
from being added
to the redundancy codebook, 
thereby reducing
the risk of
removing
informative visual tokens.

\begin{figure}[tb]
  \centering
  \includegraphics[width=0.9\linewidth]{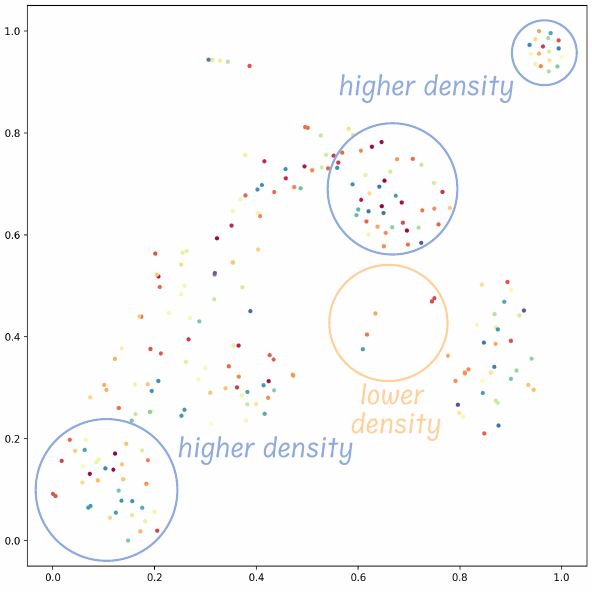}
  \caption{
    Distribution
    of redundant candidates
    in LLaVA-Next's
    feature space.
    Points of the same color indicate
    that they originate from
    the same image.
    If a cluster
    contains numerous points
    of different colors
    (\ie, has high density),
    it comprises
    a group of 
    highly similar visual tokens
    that are unlikely to
    significantly impact the
    overall information across
    different images.
  }
  \label{fig:tsne_multi_img}
\end{figure}

Upon obtaining the
redundant candidates
(lines 7 to 17
in~\Cref{alg:build_codebook}),
we further analyze
their distribution
in feature space
using t-SNE.
\Cref{fig:tsne_multi_img}
shows that
these 
redundant candidates
form clusters
with varying densities.
Clusters with higher densities
represent groups of
highly similar visual tokens
that,
when placed
in diverse
visual contexts
(\ie, different images),
do not significantly
affect the
overall information
of those images. 
Conversely, 
visual tokens
in
lower-density clusters
may have only
a negligible impact
on the overall information
of a few specific images. 
Therefore, 
we 
propose
selecting
higher-density clusters
by setting an inlier threshold $\tau_{in}$
to filter cluster sizes.

We sample 500 images
from the COCO Caption
Karpathy train split~\cite{karpathy2015deep}
to identify redundant prototypes.

\subsection{Details for Visual Token Pruning}
\label{appendix_sec:detail_for_prune}
\Cref{alg:method}
outlines the
visual token pruning process
with the
redundancy codebook
that contains
redundant prototypes
identified
using training images.
These redundant prototypes
are stored on disk
as a PyTorch Tensor
of shape $[N\times d]$.
During inference,
the redundancy codebook
is loaded onto the device,
and the only 
computational overhead
introduced by our method
is the calculation of the
similarity matrix
$\mathcal{S}^{L \times N}$
(lines 11-16 in~\Cref{alg:method}).
Once the redundancy scores
(\ie, $Sim\_max$ in line 16)
are obtained,
we directly prune
visual tokens
with redundancy scores
exceeding a predefined
threshold $r\_threshold$,
while preserving the
order of the 
remaining tokens
in the original input sequence.
Specifically, 
if an image contains
a larger number of
tokens with
high similarity 
to the redundant prototypes,
more tokens will be pruned from that image.

\textbf{The Clustering
Algorithm.}
To achieve 
satisfying
visual token clustering results,
we use the
DPC-kNN algorithm~\cite{du2016study}.
For
visual tokens
$\boldsymbol{V} = \{\boldsymbol{v}_n\}_{n=1}^{L}$,
the local density
$\rho_i$ for each
visual token
$\boldsymbol{v}_i$
is obtained by:
\begin{equation}
    \rho_i = \exp (-\frac{1}{k} \sum_{\boldsymbol{v}_j \in kNN(\boldsymbol{v}_i)} \Vert\boldsymbol{v}_j - \boldsymbol{v}_i\Vert_2),
    \label{eq:dpc_knn_1}
\end{equation}
where 
$kNN(\boldsymbol{v}_i)$
is
the $k-$nearest neighbors
of
$\boldsymbol{v}_i$
among
$\{\boldsymbol{v}_n\}_{n=1, n\neq i}^{L}$.
Next, 
the distance index 
$\delta_i$
corresponding to
$\boldsymbol{v}_i$,
\ie,
the distance between
$\boldsymbol{v}_i$
and other high-density
visual tokens
is obtained by:
\begin{equation}
    \delta_i=\left\{
    \begin{aligned}
    \min_{j:\rho_j>\rho_i}\Vert\boldsymbol{v}_j - \boldsymbol{v}_i\Vert_2 & , & \text{if }\exists \text{ } \boldsymbol{v}_j\text{ s.t. } \rho_j > \rho_i, \\
    \max_{j}\Vert\boldsymbol{v}_j - \boldsymbol{v}_i\Vert_2 & , & \text{otherwise}.
    \end{aligned}
    \right.
    \label{eq:dpc_knn_2}
\end{equation}
Subsequently,
visual tokens
with relatively high
$\rho \times \delta$ values
are identified as cluster centers,
and other visual tokens
are assigned to their
nearest cluster center
based on
Euclidean distance.
The cluster size
represents the number
of visual tokens
within each cluster.
In the \textit{single-token-input} experiment,
the hyper-parameter $k$
is set to 16.
In the  \textit{cascaded leave-one-out}
experiment, $k$ is set to 64
for LLaVA-1.5 and LLaVA-Next,
and 24 for LLaVA-OneVision.

\begin{algorithm}[tb]
\caption{PyTorch style pseudocode for
visual token pruning
with
the \textit{redundancy codebook}
during inference}
\begin{algorithmic}[1]
\STATE import torch
\STATE import torch.nn.functional as F
\STATE \textcolor{darkgreen}{\textit{\# load the 
\textit{redundancy codebook}, [N, d]}}
\STATE $codebook$ = torch.load($codebook\_path$).to($device$)
\STATE $r\_threshold$ = 0.5
\STATE \textcolor{darkgreen}{\textit{\# get image features without ViT$-[cls]$ token}}
\STATE $image\_feats$ = vision\_tower($images$)[:, 1:, :]
\STATE $image\_feats$ = mm\_projector($image\_feats$)
\STATE $bs$, $L$, $d$ = $image\_feats$.shape
\STATE \textcolor{darkgreen}{\textit{\# calculate the similarity matrix $\mathcal{S}^{L \times N}$}}
\STATE $i\_norm$ = F.normalize($image\_feats$, p=2, dim=-1)
\STATE $cb\_norm$ = F.normalize($codebook$, p=2, dim=-1)
\STATE $cb\_norm$ = $cb\_norm$.unsqueeze(0).repeat($bs$, 1, 1)
\STATE $cb\_norm$ = $cb\_norm$.transpose(1, 2)
\STATE $Sim$ = torch.matmul($i\_norm$, $cb\_norm$)
\STATE $Sim\_max$, $\_$ = torch.max($Sim$, dim=-1)
\STATE \textcolor{darkgreen}{\textit{\# prune visual tokens}}
\STATE $indices$ = $Sim\_max <= r\_threshold$
\STATE $selected\_i$ = [$image\_feats$[i][$indices$[i]] \textbf{for} $i$ \textbf{in} range($bs$)]
\end{algorithmic}
\label{alg:method}
\end{algorithm}


\section{Experiments}

\begin{figure}[tb]
  \centering
  \includegraphics[width=0.9\linewidth]{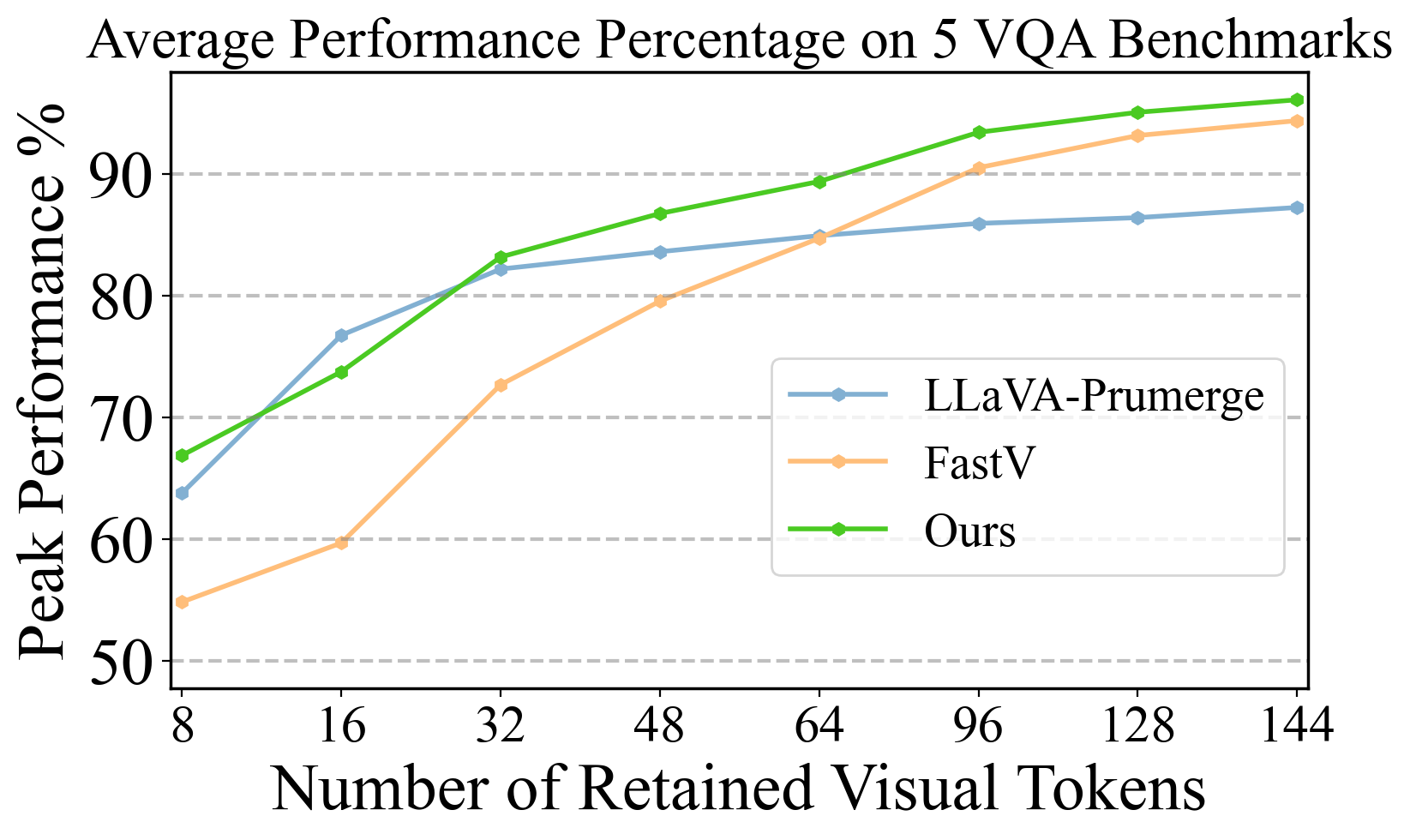}
  \caption{
    Detailed performance comparison
    between our proposed
    visual token pruning
    method and 
    existing methods that depend on
    MLLMs' 
    \textit{intermediate
    activation states}.
    At extreme
    visual token pruning rates
    (75\% to 99\%),
    our method achieves
    the best overall performance
    on five single 
    image VQA benchmarks.
  }
  \label{fig:fine_granularity_compare}
\end{figure}

\subsection{Detailed Experiment Setting}
\label{appendix_sec:detailed_exp_settings}
We follow
the official evaluation toolkits
for single image VQA,
(MME, POPE, SEED, MMBench, and RealWorldQA),
multi-image VQA
(Mantis-test and MUIRBench)
and video QA (MVBench).
For the POPE benchmark,
we report the averaged results
across the 
\textit{random},
\textit{popular},
and \textit{adversarial} subsets.
For image captioning,
we use the \textit{pycocoevalcap} package
to compute quantitative metrics.

\textbf{Controlling the Number of Visual Tokens.}
During inference, the number of retained 
visual tokens $R$ is controlled by 
the $r\_threshold$ parameter
in~\Cref{alg:method}.
To ensure a fair comparison
under the same visual token budget,
we adjust $r\_threshold$
so that
the average number of
visual tokens
across all
test samples
is the same as
that of
LLaVA-Prumerge and FastV.


\textbf{Hyperparameters.}
For LLaVA-1.5 and LLaVA-Next,
the thresholds $\tau_{prob}$,
$\tau_{out}$,
$\tau_{in}$,
and
$\tau_{jsd}$
are set to
0.1, 8, 64, and 2e-3,
respectively.
For LLaVA-OneVision,
these thresholds are set to
0.08, 3, 16 and 1.5e-3, respectively.
These hyperparameter configurations
result in
454, 969, and 310
redundant prototypes
for LLaVA-1.5, 
LLaVA-Next, 
and LLaVA-OneVision,
respectively.
In the 
\textit{single-input}
experiment,
the \textit{top-1 probability}
is computed among the
top 50 ranked candidates.
in the 
\textit{cascaded leave-one-out}
experiment,
the JSD value
is calculated among the
top 20 ranked candidates.
The results of
the random baseline
are obtained from
three independent runs
with different random seeds.

\textbf{Environment.}
All experiments
are conducted on NVIDIA 3090-24G GPUS.
Our method is implemented
with PyTorch and the Huggingface
Transformers Library.

\begin{figure}[h]
  \centering
  \includegraphics[width=0.88\linewidth]{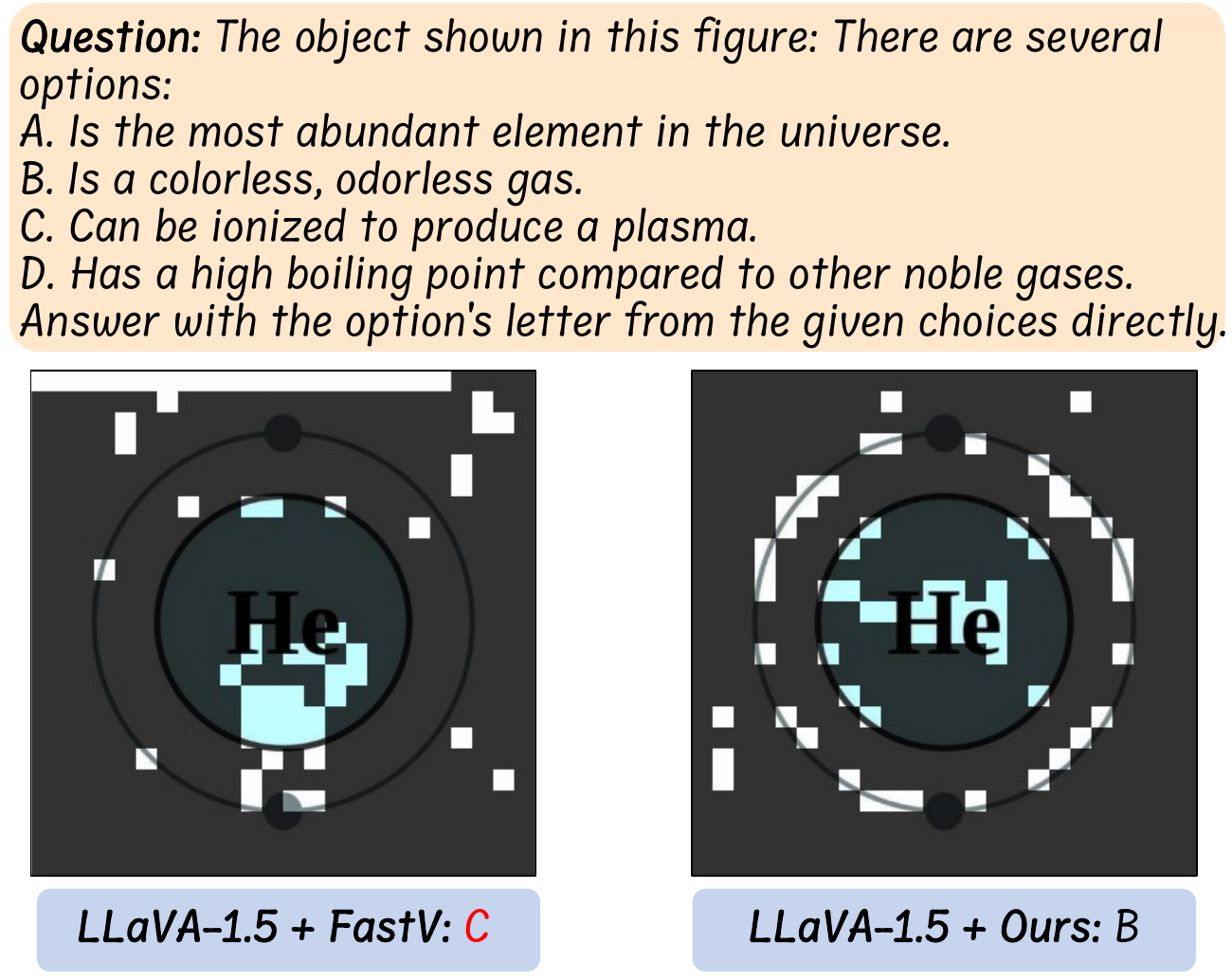}
  \caption{
    Qualitative examples
    on the MMB benchmark.
    For challenging questions,
    our proposed method
    effectively
    prunes redundant
    patches
    and retain
    the important
    image regions
    that show
    the chemical element $H_e$.
    Wrong answer
    is highlighted in red.
  }
  \label{fig:qualitative_mmb}
\end{figure}

\subsection{Comparison with
Existing Methods}
\label{appendix_sec:perf_comparison}
In this work,
we propose to identify
redundant visual tokens
by investigating the
direct impact
of each visual token
on MLLMs' output.
We compare our proposed method
with
visual-centric and
instruction-based
approaches,
which assess visual redundancy
based on MLLMs'
intermediate activation states.
For the visual-centric approach,
we compare with a representative method
LLaVA-Prumerge,
which prunes visual tokens
that exhibit lower correlation
with the ViT$-[cls]$ token.
For the instruction-based approach,
we compare with 
FastV,
which prunes visual tokens
with lower text-to-image attention scores
in the language decoder
(where the query corresponds to the last token
in the input sequence).
We use their official
code implementations
and default hyperparameters.
To ensure a fair comparison,
we adhere to a
training-free
visual token pruning setting,
removing visual tokens
before sending them to the LLM.
For FastV,
we follow its default setting
of pruning visual tokens
at the LLM's second layer.

\begin{figure*}[h]
  \centering
  \includegraphics[width=0.88\linewidth]{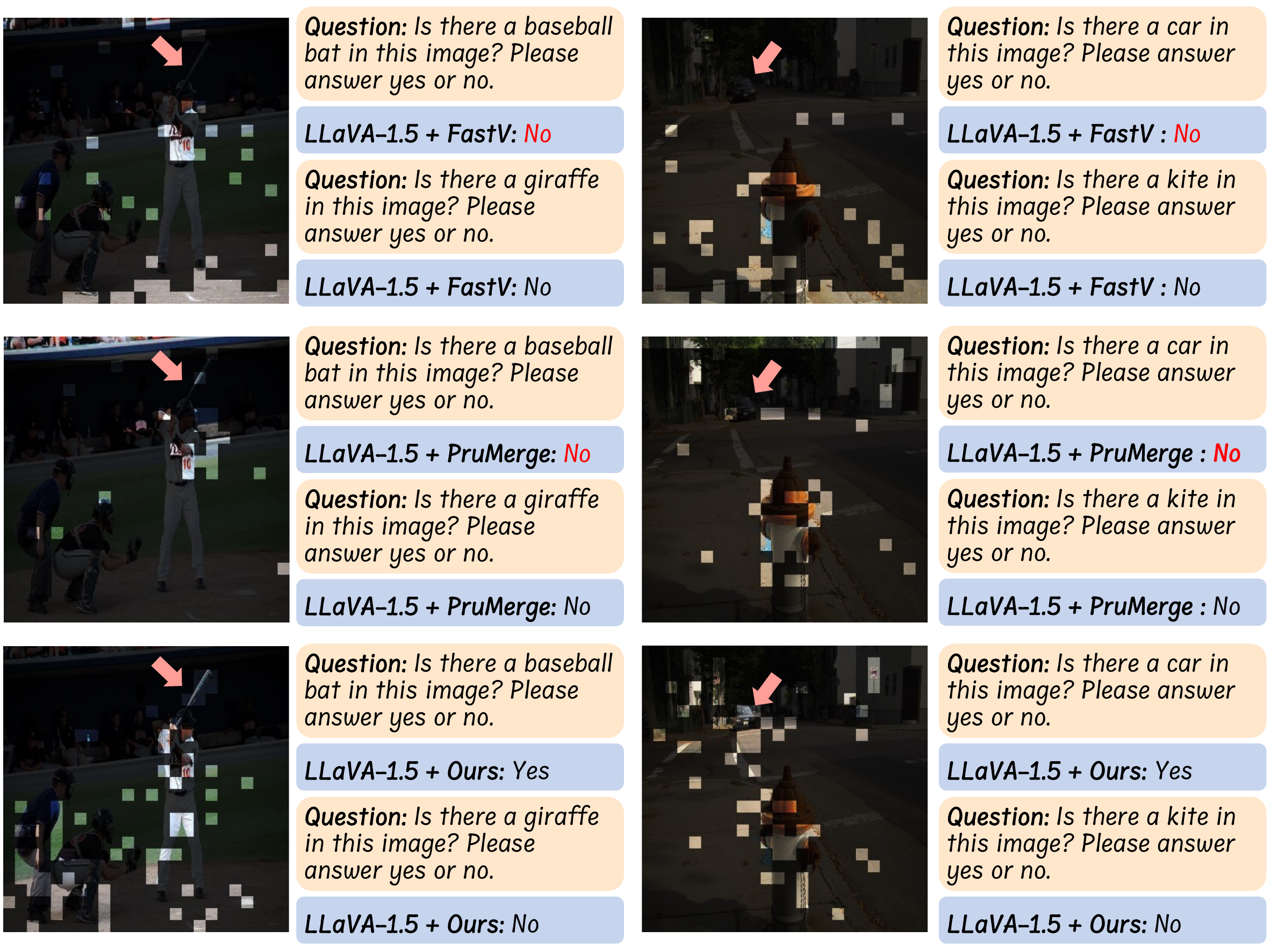}
  \caption{Qualitative examples
    on the MME benchmark.
    Our proposed method
    effectively allocates
    the limited visual token budget
    to critical visual elements
    in the images
    (indicated by the pink arrow).
    Wrong answers
    are highlighted in red.
  }
  \label{fig:qualitative_mme}
\end{figure*}


To facilitate
a more detailed
performance comparison
with existing methods, 
we further examine
the performance of
LLaVA-1.5
across an extreme range of
visual token removal 
(75\%–99\%). 
\Cref{fig:fine_granularity_compare}
illustrates
the average performance
of our proposed method, 
LLaVA-Prumerge, 
and FastV
across five 
single-image VQA benchmarks
(MME, POPE, MMBench, SEED and RealWorldQA).
These results are
presented as percentages
relative to 
the peak performance
achieved when all visual tokens
(576 tokens) are
included.
We observe that
FastV's performance
rapidly deteriorates
when the number of
input visual tokens
falls below 96.
On the other hand,
LLaVA-Prumerge
exhibits robust performance
when retaining
a minimal number of
visual tokens
(8 to 32 tokens), 
but its performance
is significantly worse
than FastV
when retaining
more
visual tokens
(96 to 144).
Overall,
our proposed method
achieves
the highest performance
across the entire token removal range,
markedly 
outperforming
LLaVA-Prumerge
and 
FastV.

Qualitative
examples
of our proposed method
and existing approaches
are shown in~\Cref{fig:qualitative_mmb}
and~\Cref{fig:qualitative_mme}.
For both
text-rich images
and natural photographs,
our method
effectively allocates
the 
limited visual token budget
to critical visual elements
within the images,
such as
the region presenting
the chemical element $H_e$
in~\Cref{fig:qualitative_mmb},
and the areas containing the
baseball bat, 
baseball players, 
as well as 
the car,
fire hydrant, 
road signs,
and background buildings
in~\Cref{fig:qualitative_mme}.
This advantage
enables 
our method to
accurately address
a variety of
challenging questions
while retaining
a minimal number of
visual tokens, 
thereby outperforming
existing methods.

\begin{figure*}[tb]
  \centering
  \includegraphics[width=0.88\linewidth]{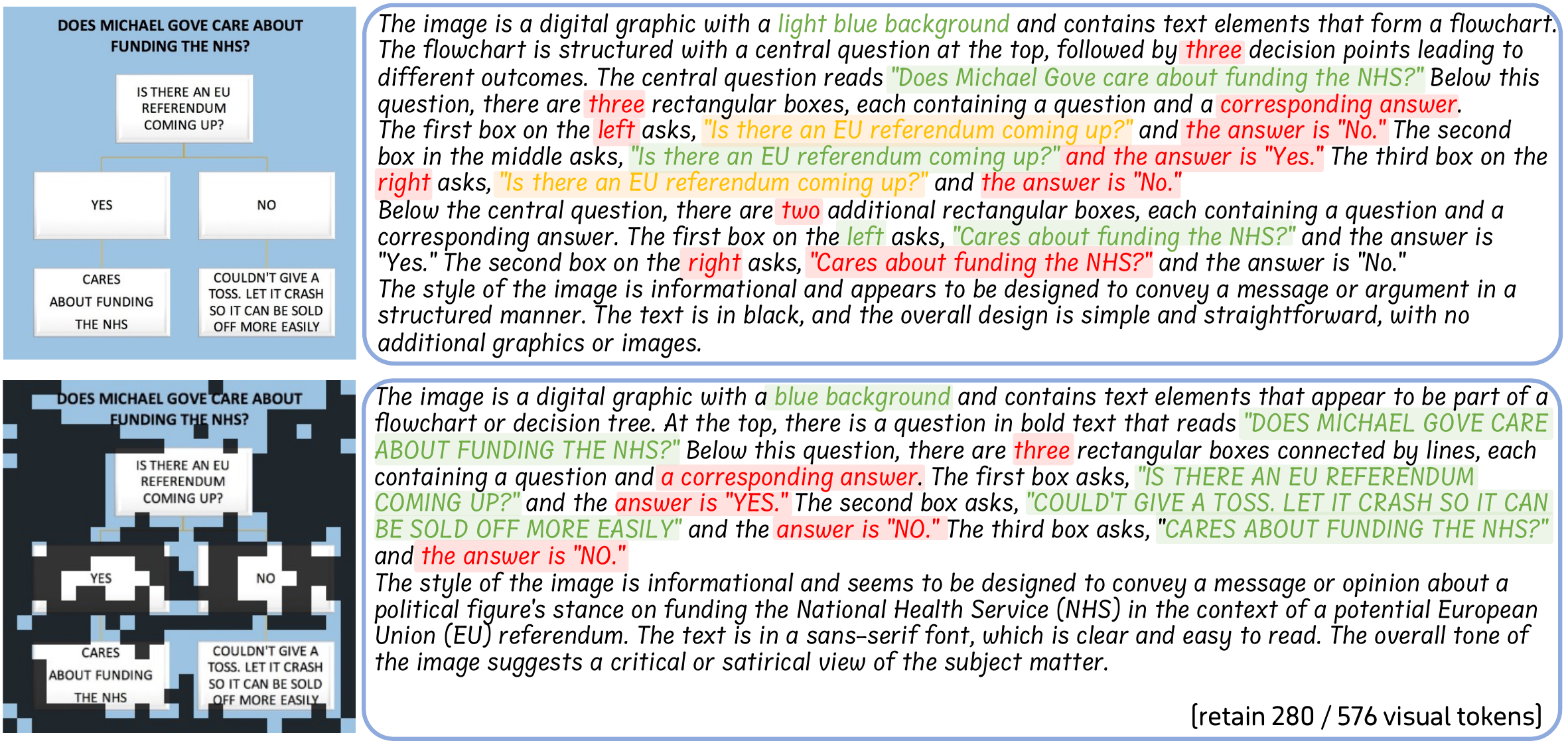}
  \caption{
    Image detailed captioning
    results on text-rich images.
    Our method
    eliminates redundant
    single-color background patches
    and retains
    key elements in the flow chart,
    resulting in
    less errors
    in the generated description.
    The correct content
    in the image
    is highlighted in green, 
    ambiguous or repetitive content
    in yellow,
    and incorrect content in red.
  }
  \label{fig:dc100_part2}
\end{figure*}

\subsection{Experiments on 
More Challenging Tasks}
We further validate
the effectiveness
of our proposed method
on more challenging
vision-language tasks,
including
detailed image captioning
and spotting subtle differences
between two images.

\begin{figure*}[tb]
  \centering
  \includegraphics[width=0.88\linewidth]{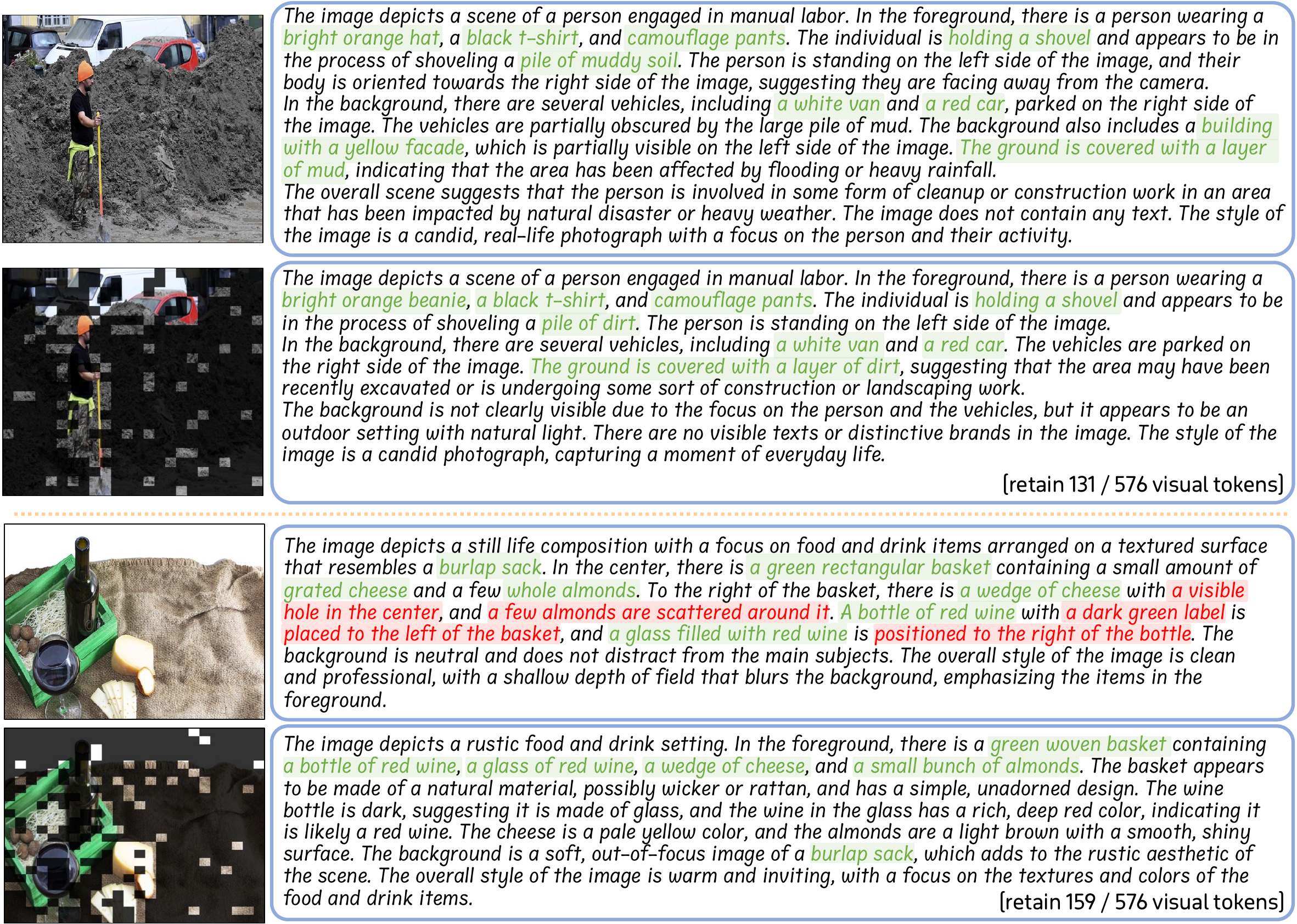}
  \caption{
    Image detailed captioning
    results
    on natural photographs.
    Our proposed method
    maintains almost the 
    same level of detail
    even after
    removing 
    three-quarters
    of the input visual tokens.
    Moreover,
    removing 
    redundant visual tokens
    may help reduce
    hallucinatory content
    in image descriptions.
    Correct and erroneous
    content are highlighted
    in green and red color,
    respectively.
  }
  \label{fig:dc100_part1}
\end{figure*}

\begin{figure*}[tb]
  \centering
  \includegraphics[width=0.9\linewidth]{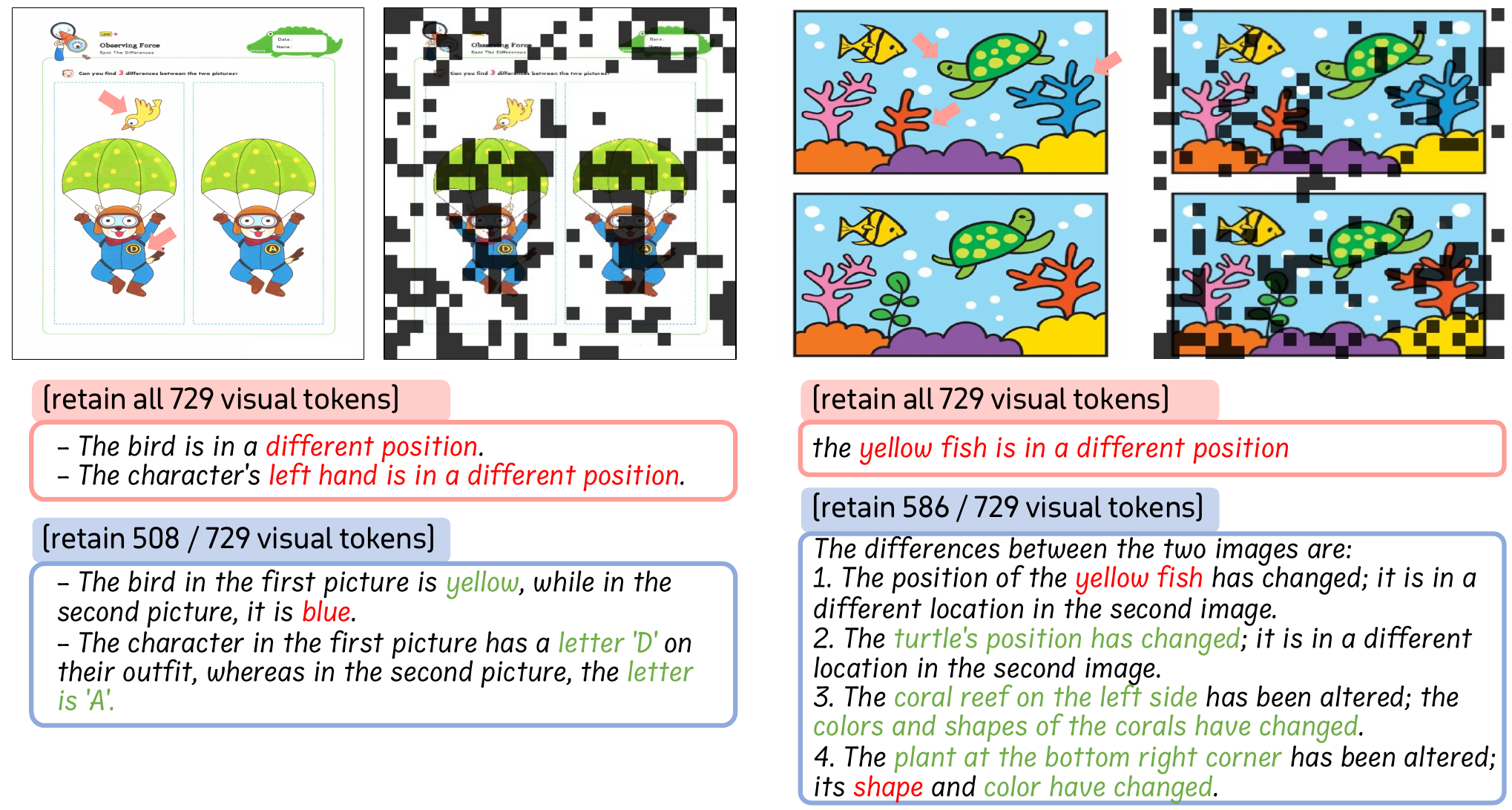}
  \caption{
    Spot-the-difference
    results
    from the LLaVA-OneVision model.
    These examples
    demonstrate that
    the model's ability
    to discern 
    fine-grained
    distinctions
    is enhanced 
    when a subset of
    redundant visual tokens
    is removed
    based on our proposed method.
    Correct and erroneous
    descriptions are highlighted
    in green and red color,
    respectively.
  }
  \label{fig:spot_the_diff}
\end{figure*}

\subsubsection{Detailed Image Captioning}

We further
present 
examples
from the
image Detailed Caption
(Image-DC) 
test 
set\footnote{\url{https://huggingface.co/datasets/lmms-lab/DC100_EN}},
where MLLMs are instructed to
describe the image in detail, 
covering the
attributes of objects, scenes and background.
Qualitative results
are shown in~\Cref{fig:dc100_part1}
and~\Cref{fig:dc100_part2}.
For images
containing 
multiple objects, 
our approach
maintains
a high level of detail
in the descriptions,
even after
removing three-quarters
of the input visual tokens.
As illustrated
in~\Cref{fig:dc100_part1},
our approach 
ensures that
the key descriptions
are preserved,
such as the workers' attire
(orange beanie, 
black t-shirt, 
camouflage pants),
actions
(shoveling dirt),
and 
background objects
(a white van and a red car),
while removing
redundant image patches
that merely depict soil. 
Additionally,
we observe that
the pruning
redundant visual tokens
may
help
mitigate
visual hallucinations
in MLLMs.
For example,
\Cref{fig:dc100_part1}
demonstrates that
eliminating about
three-quarters of the
input visual tokens
(which only
display white backgrounds
and fabric textures)
reduces the
hallucinatory content
in the textual response
(\eg, a wedge of 
cheese with a hole
in the center
and a dark green label
on the left side of the basket).
Additionally,
\Cref{fig:dc100_part2}
illustrates that
our method
generalizes well
to text-rich images,
as it
preserves all
textual content
in the flowchart
while eliminating
single-color background patches.
We find that
removing
these background patches
reduces both
the omission of 
critical information
and errors
in the model's
description.
In summary,
our method
effectively
prunes redundant image patches
across various domains
and has the potential to
improve
the accuracy of
visual comprehension
for MLLMs.

\subsubsection{Spot Subtle Differences between Images}

We further assess
the ability
of our method
to help MLLMs
recognize image details.
To this end,
we collect spot-the-difference game images
from copyright-free websites
and 
instruct the LLaVA-OneVision model
to identify discrepancies
between them.
In this experiment,
the input images are not divided into sub-images.
As shown in~\Cref{fig:spot_the_diff},
when all 729
visual tokens
are provided
as input,
the model
struggles to identify 
fine-grained differences
between the images
and frequently generates hallucinations
(\eg, the yellow fish,
the bird, and the
character's hand in a
different position). 
However,
by removing
a portion of
the input visual tokens,
the MLLM is better able to
identify the differences
(\eg, the letter on the character's outfit
and the turtle's position)
and describe them
more precisely
(\eg, changes to the coral reef).
We thus hypothesize that
redundant visual tokens
may obscure
fine-grained visual details,
and 
removing them
enables the MLLM to
more effectively recognize
these details.

The prompt for this task is:
\textit{The user is playing a spot-the-difference game.
The provided image displays two pictures arranged either vertically or horizontally. Please help the user identify all the differences between the two images.
Please provide accurate answers in a bullet list format.}


\subsubsection{Summary}

In summary, 
our experiments 
on
image detailed captioning
and 
spot-the-difference
tasks
demonstrate that
redundant visual tokens
can obscure visual details in images.
Eliminating
these redundancies
has the potential to
improve MLLMs' accuracy
in visual understanding.
We hope that
our methods
for visual redundancy analysis
and visual token pruning,
along with our experimental results,
will inspire
future research
into the
visual understanding 
behaviors of MLLMs.

%
%

\end{document}